\documentclass[lettersize,journal]{IEEEtran}
\usepackage{amsmath,amsfonts}
\usepackage{amssymb}
\usepackage{mathtools}
\usepackage{algorithmic}
\usepackage{algorithm}
\usepackage{array}
\usepackage[caption=false,font=normalsize,labelfont=sf,textfont=sf]{subfig}
\usepackage{textcomp}
\usepackage{stfloats}
\usepackage{url}
\usepackage{verbatim}
\usepackage{graphicx}
\usepackage{cite}
\usepackage{upquote}
\usepackage{bbm}
\usepackage{multirow}
\usepackage{todonotes}

\newtheorem{theorem}{Theorem}[section]
\newtheorem{definition}[theorem]{Definition}
\newtheorem{problem}{\textbf{Problem}}

\newcommand{\subparagraph}{}
\DeclareMathOperator*{\argmax}{argmax}

\DeclarePairedDelimiter\ceil{\lceil}{\rceil}

\begin{document}

\title{Heuristic Satisficing Inferential Decision Making in Human and Robot Active Perception}

\author{Yucheng~Chen,
        Pingping~Zhu,~\IEEEmembership{Member,~IEEE}
        Anthony Alers,
        Tobias Egner,
        Marc A. Sommer,
        Silvia Ferrari,~\IEEEmembership{Senior Member,~IEEE}
\thanks{Y. Chen (yc2383@cornell.edu), and S. Ferrari (ferrari@cornell.edu)
are with the Sibley School of Mechanical and Aerospace Engineering at Cornell
University, Ithaca, NY, 14850; P. Zhu (zhup@marshall.edu) is with the College of Engineering
and Computer Sciences at Marshall University, Huntington, WV, 25755; A. Alers
(anthony.alers@duke.edu), T. Egner (tobias.egner@duke.edu) and M. Sommer
(marc.sommer@duke.edu) are with Duke University, Durham, NC, 27708.
David Zielinski at Duke University is acknowledged for assistance with the human experiments.}}

\markboth{Journal of \LaTeX\ Class Files,~Vol.~14, No.~8, August~2021}%
{Shell \MakeLowercase{\textit{et al.}}: A Sample Article Using IEEEtran.cls for IEEE Journals}

\maketitle

\begin{abstract}
Inferential decision-making algorithms typically assume that
an underlying probabilistic model of decision alternatives and outcomes may
be learned \emph{a priori} or online. Furthermore, when applied to robots in 
real-world settings they often perform unsatisfactorily or fail to accomplish the 
necessary tasks because this assumption is violated and/or they experience 
unanticipated external pressures and constraints. Cognitive studies presented 
in this and other papers show that humans cope with complex and unknown 
settings by modulating between near-optimal and satisficing solutions, including 
heuristics, by leveraging information value of available environmental cues that
are possibly redundant. Using the benchmark inferential decision problem known
as ``treasure hunt", this paper develops a general approach for investigating and
modeling active perception solutions under pressure. By simulating treasure hunt 
problems in virtual worlds, our approach learns generalizable strategies from high 
performers that, when applied to robots, allow them to modulate between optimal 
and heuristic solutions on the basis of external pressures and probabilistic models, 
if and when available. The result is a suite of active perception algorithms for 
camera-equipped robots that outperform treasure-hunt solutions obtained via cell 
decomposition, information roadmap, and information potential algorithms, in both 
high-fidelity numerical simulations and physical experiments. The effectiveness of 
the new active perception strategies is demonstrated under a broad range of 
unanticipated conditions that cause existing algorithms to fail to complete the 
search for treasures, such as unmodelled time constraints, resource constraints, 
and adverse weather (fog).
\end{abstract}


\section{Introduction}
\label{sec:Introduction}

\IEEEPARstart{R}{ational} inferential decision-making theories in both human and
autonomous robot studies assume knowledge of a world model, such that near-optimal or
satisficing strategies may be achieved by maximizing an appropriate utility function
and/or satisficing mathematical constraints \cite{SimonBehavioral55, SimonRational79,
	CaplinBasic14,NicolaidesLimits88,SimonSciences19}.
When a probabilistic world model is available, either because it is learned online
or \emph{a priori}, a variety of approaches, such as optimal control, robot/sensor planning,
and maximum utility theories may be applied to inferential decision-making problems
for robot active perception, planning, and feedback control \cite{FishburnSubjective81,
	LebedevCortical05,ScottOptimal04,TodorovOptimal02, FerrarInformation21,
	LatombeRobotMotionPlanningBook12, LavallePlanningAlgorithmsBook06}.
Many ``model-free" reinforcement learning (RL) and approximate dynamic programming
(ADP) approaches have also been developed on the basis of the assumption that a partial
or imperfect model is available in order to predict the next system state and/or ``cost-to-go",
and optimize the immediate and potential future rewards, such as information value
\cite{BertsekasDynamicProgramming12, SiADPhandbook04, PowellApproximateDP07,
	FerrariInformation09, SuttonReinforcement18, WieringReinforcement12}.

Humans have also been shown to use internal world models for inferential decision-making
whenever possible, a characteristic first referred to as ``substantial rationality"
in \cite{SimonRational79,SimonBehavioral55}. As also shown by the human studies
on passive and active satisficing perception presented in this paper, given sufficient data,
time, and informational resources, a globally rational human decision-maker uses an
internal model of available alternatives, probabilities, and decision consequences to optimize
both decision and information value in what is known as a ``small-world" paradigm \cite{SavageFoundations72}.
In contrast, in ``large-world" scenarios, decision-makers face environmental pressures that
prevent them from building an internal model or quantifying rewards, because of pressures such
as missing data, time and computational power constraints, or sensory deprivation, yet still
manage to complete tasks by using ``bounded rationality" \cite{SimonModels97}.
Under these circumstances, optimization-based methods may not only be infeasible, returning
no solution, but also cause disasters resulting from failing to take action \cite{GigerenzerHeuristic11}.
Furthermore, Simon and other psychologists have shown that humans can overcome these limitations
in real life via ``satisficing decisions"  that modulate between near-optimal strategies and
the use of heuristics to gather new information and arrive at fast and ``good-enough"
solutions to complete relevant tasks.

To develop satisficing solutions for active robot perception, herein, we consider here the
class of sensing problems known as treasure hunt  \cite{CaiCellDecomposition09,ZhangInformation09,
	FerrariInformation09, ZhangComparisonInfoFunc11}. The mathematical model of the problem, comprised of
geometric and Bayesian network descriptions demonstrated in \cite{CaiCellDecomposition09,
	FerrarInformation21}, is used to develop a new experimental design approach that
ensures humans and robots experience the same distribution of treasure hunts in
any given class, including time, cost, and environmental pressures inducing satisficing
strategies. This novel approach enables not only the readily comparison of the human-robot
performance but also the generalization of the learned strategies to any treasure hunt
problem and robotic platform. Hence, satisficing strategies are modeled using human
decision data obtained from passive and active satisficing experiments, ranging from
desktop to virtual reality human studies sampled from the treasure hunt model. Subsequently,
the new strategies are demonstrated through both simulated and physical experiments
involving robots under time and cost pressures, or subject to sensory deprivation (fog).

\begin{figure}
     \centering
    \subfloat[\label{subfig:DiVEClear}]{
        \includegraphics[width=0.22\textwidth]{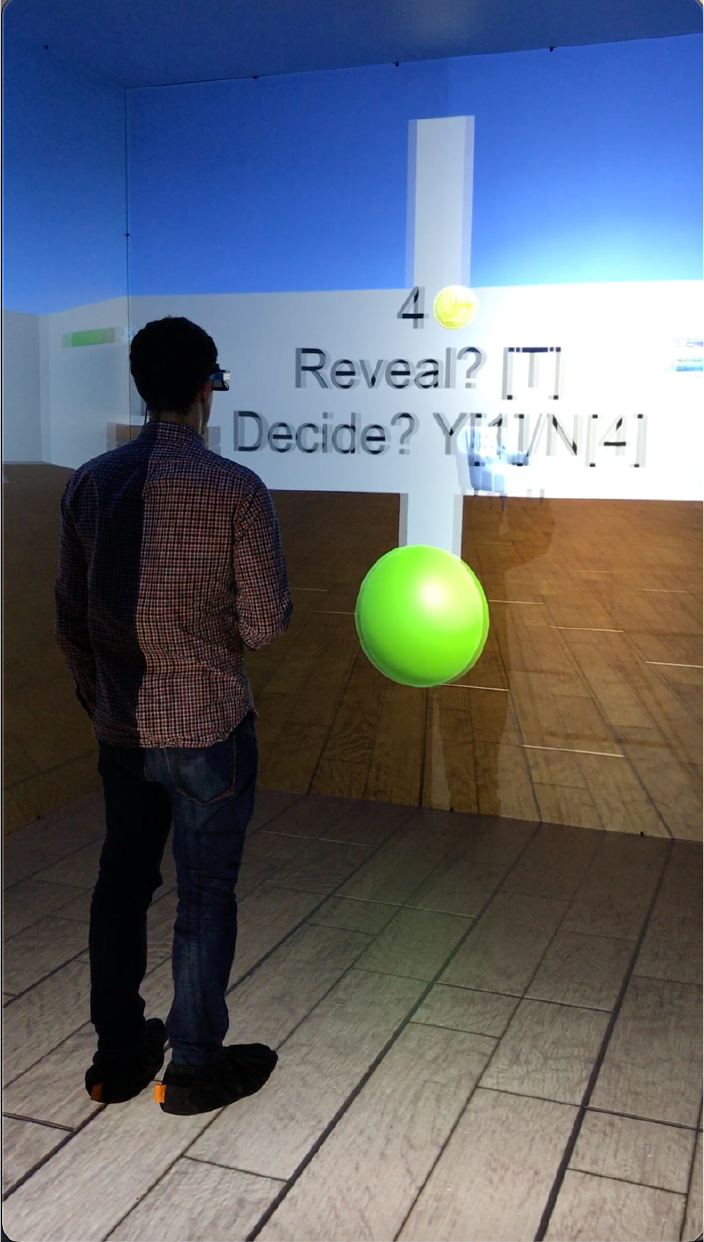}
    }
    \hfill
    \subfloat[\label{subfig:DiVEFog}]{
        \includegraphics[width=0.22\textwidth]{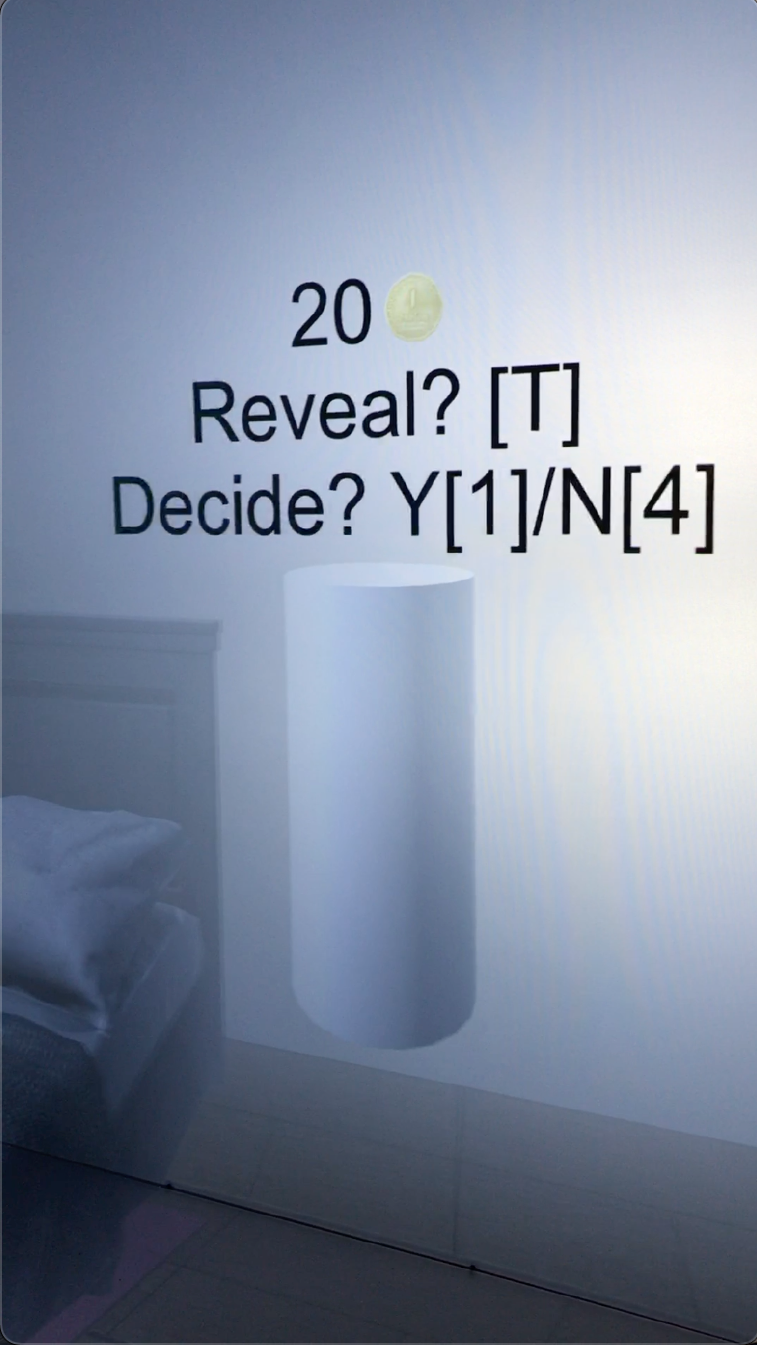}
    }
    \caption{Human participant solving treasure hunt problem under no pressures (a),
    	and under sensory deprivation (fog) (b) in the Duke immersive Virtual Environment \cite{DukeDiVE22}.}
    \label{fig:HumanDiVE}
\end{figure}

The treasure hunt problem under pressure, formulated in Section \ref{sec:Probformulation}
and referred to as \textit{satisficing treasure hunt} herein, is an extension of the robot
treasure hunt presented in \cite{CaiCellDecomposition09,ZhangInformation09},
which introduces motion planning and inference in the search for Spanish treasures
originally used in \cite{SimonOptimal75} to investigate satisficing decisions in humans.
Whereas the search for Spanish treasures amounts to searching a (static) decision tree with
hidden variables, the robot treasure hunt involves a sensor-equipped robot searching for
targets in an obstacle-populated workspace. As shown in \cite{FerrarInformation21}
and references therein, the robot treasure hunt paradigm is useful in many mobile sensing
applications involving multi-target detection and classification. In particular, the problem
highlights the coupling of action decisions that change the physical state of the robot
(or decision-maker) with test decisions that allow the robot to gather information from the
targets via onboard sensors. In this paper, the satisficing treasure hunt is introduced
to investigate and model human satisficing perception strategies under external pressures in
passive and active tasks, first via desktop simulations and then in the Duke immersive Virtual
Environment (DiVE) \cite{DukeDiVE22}, as shown in Fig. \ref{fig:HumanDiVE}.

To date, substantial research has been devoted to solving treasure hunt problems for many
robots/sensor types, in applications as diverse as demining infrared sensors and underwater
acoustics, under the aforementioned ``small-world" assumptions \cite{FerrarInformation21}.
Optimal control and computational geometry solution approaches, such as cell decomposition
\cite{CaiCellDecomposition09}, disjunctive programming \cite{SwinglerDuality13},
and information roadmap methods (IRM) \cite{ZhangInformation09}, have been developed
for optimizing robot performance by minimizing the cost of traveling through the workspace
and processing sensor measurements, while maximizing the sensor rewards such as information
gain. All these existing methods assume prior knowledge of sensor performance and of the
workspace, and are applicable when the time and energy allotted to the robot are adequate
for completing the sensing task. Information-driven path planning algorithm integrated
with online mapping, developed in \cite{ZhuScalable19,LiuGround19,GeSimultaneous11},
have extended former treasure hunt solutions to problems in which a prior model of the workspace
is not available and must be obtained online. Optimization-based algorithms have also been
developed for fixed end-time problems with partial knowledge of the workspace, on the basis
of the assumption that a probabilistic model of the information states and unlimited sensor
measurements are available \cite{RosselloInformation21}.  This paper builds on this
previous work to develop heuristic strategies applicable when uncertainties cannot be
learned or mathematically modeled in closed form, and the presence of external pressures
might prevent task completion, e.g., adverse weather or insufficient time/energy.

Inspired by previous findings on human satisficing heuristic strategies 
\cite{GigerenzerHeuristic11,GigerenzerHeuristicCognitivePsychology91, 
	GigerenzerReasoningBoundedRationality96,GigerenzerGut07,OhSatisficing16},
this paper develops, implements, and compares the performance between
existing treasure hunt algorithms and human participants engaged in the same
sensing tasks and experimental conditions by using a new design approach. Subsequently,
human strategies and heuristics outperforming existing state-of-the-art algorithms are
identified and modeled from data in a manner that can be extended to any sensor-equipped
autonomous robot. The effectiveness of these strategies is then demonstrated with camera-equipped
robots via high-fidelity simulations as well as physical laboratory experiments. In particular,
human heuristics are modeled by using the ``three building blocks" structure for formalizing
general inferential heuristic strategies presented in \cite{GigerenzerSimple99}.
The mathematical properties of heuristics characterized by this approach are then
compared with logic and statistics, according to the rationale in \cite{GigerenzerHeuristic11}.

Three main classes of human heuristics for inferential decisions exist:
recognition-based decision-making \cite{RatcliffSimilarity89,GoldsteinModels02},
one-reason decision-making \cite{NewellTake03,GigerenzerGut07},
and trade-off heuristics\cite{LichtmanKeys08}. Although categorized by
respective decision mechanisms, these classes of human heuristics have been
investigated in disparate satisficing settings, thus complicating the determination of
which strategies are best equipped to handle different environmental pressures.
Furthermore, existing human studies are typically confined to desktop simulations
and do not account for action decisions pertaining to physical motion and path
planning in complex workspaces. Therefore, this paper presents a new experimental
design approach (Section \ref{sec:Humanstudies}) and tests in human participants to
analyze and model satisficing active perception strategies (Section \ref{sec:Activestratgymodeling})
that are generalizable and applicable to robot applications, as shown in Section \ref{sec:RoboApp}.

The paper also presents new analysis and modeling studies of human satisficing strategies 
in both passive and active perception and decision-making tasks (Section \ref{sec:Humanstudies}). 
For passive tasks, time pressure on inference is introduced to examine subsequent effects on 
human decision-making in terms of decision model complexity and information gain. 
The resulting heuristic strategies (Section \ref{sec:Passivestrategymodeling}) extracted 
from human data demonstrate adaptability to varying time pressure, thus enabling inferential 
decision-making to meet decision deadlines. These heuristics significantly reduce the 
complexity of target feature search from an exhaustive search $O(2^n)$ to $O(nlog(n) + n)$, 
where $n$ is the number of target features. Additionally, they exhibit superior classification 
performance when compared to optimizing strategies that utilize all target features for 
inference (Section \ref{sec:Passiveperformeval}), demonstrating the less-can-be-more 
effect \cite{GigerenzerHeuristic11}.

For active tasks, when the sensing capabilities are significantly hindered, such as in 
adverse weather conditions, human strategies are found to amount to highly effective 
heuristics that can be modeled as shown in Section \ref{sec:Activestratgymodeling}, 
and generalized to robots as shown in Section \ref{sec:RoboApp}. The human strategies 
discovered from human studies are implemented on autonomous robots equipped with vision
sensors and compared with existing planning methods (Section \ref{sec:RoboApp}) through
simulations and physical experiments in which optimizing strategies fail to complete the task 
or exhibit very poor performance. Under information cost pressure, a decision-making 
strategy developed using mixed integer nonlinear program (MINLP) \cite{CaiCellDecomposition09, 
	ZhangInformation09} was found to outperform existing solutions as well as human strategies
(Section \ref{sec:RoboApp}). By complementing the aforementioned heuristics, the MINLP 
optimizing strategies provide a toolbox for active robot perception under pressures that is 
verified both in experiments and simulations.

\section{Treasure Hunt Problem Formulation}
\label{sec:Probformulation}

This paper considers the active perception problem known as treasure hunt, in which a 
mobile information-gathering agent, such as a human or an autonomous robot, must find 
and localize all important targets, referred to as \emph{treasures}, in an unknown workspace 
$\mathcal{W} \subset \mathbb{R}^3$. The number of possible treasures or \emph{targets}, $r$, 
is unknown \emph{a priori}, and each target $i$ may constitute a treasure or another object, 
such as a clutter or false alarm, such that its classification may be represented by a random 
and discrete hypothesis variable $Y_i$ with finite range $\mathcal{Y} = \{y_j ~|~ j \in {\mathcal{J}}\}$, 
where $y_j$ represents the $j$th category of $Y_i$. While $Y_i$ is hidden or non-observable, 
it may be inferred from $\mathit{p}_i \in \mathbb{Z}$ observed features among a set of $n$ 
discrete random variables $X_i = \{X_{i,1}, \ldots, X_{i,n}\}$, and the $l$th ($1 \leq l \leq n$) 
feature has a finite range $\mathcal{X}_l = \{x_{l,j} ~|~ j \in \mathcal{N}\}$ 
(see \cite{FerrarDemining06, ZhangComparisonInfoFunc11, FerrariInformation09} for more details). 
At the onset of the search, $X_i$ and $Y_i$ are assumed unknown for all targets, as are the 
number of targets and treasures present in $\mathcal{W}$. Thus, the agent must first navigate 
the workspace to find the targets and, then, observe their features to infer their classification.

All $r$ targets are fixed, at unknown positions
$\mathbf{x}_1,...,\mathbf{x}_r \in \mathcal{W}$, and must be detected, observed, and classified 
using onboard sensors with bounded field-of-view (FOV) \cite{FerrarInformation21}:
\begin{definition}[Field-of-view (FOV)]
	\label{def:FOV} For a sensor characterized by a dynamic state, in a workspace 
	$\mathcal{W} \subset \mathbb{R}^3$, the FOV is defined as a closed and bounded subset 
	$\mathcal{S} \subset \mathcal{W}$ such that a target feature $X_{i,l}$ may be observed 
	at any point $\mathbf{x}_i \in \mathcal{S}$.
\end{definition}

In order to obtain generalizable strategies for camera-equipped robots, in both human and robot 
studies knowledge of the targets is acquired, at a cost, through vision, and the sensing process is 
modeled by a probabilistic Bayesian network learned from data \cite{FerrarInformation21}.

\begin{figure*}
	\centering
	\subfloat[\label{subfig:HumanFOV}]{
		\includegraphics[width = 0.45\textwidth]{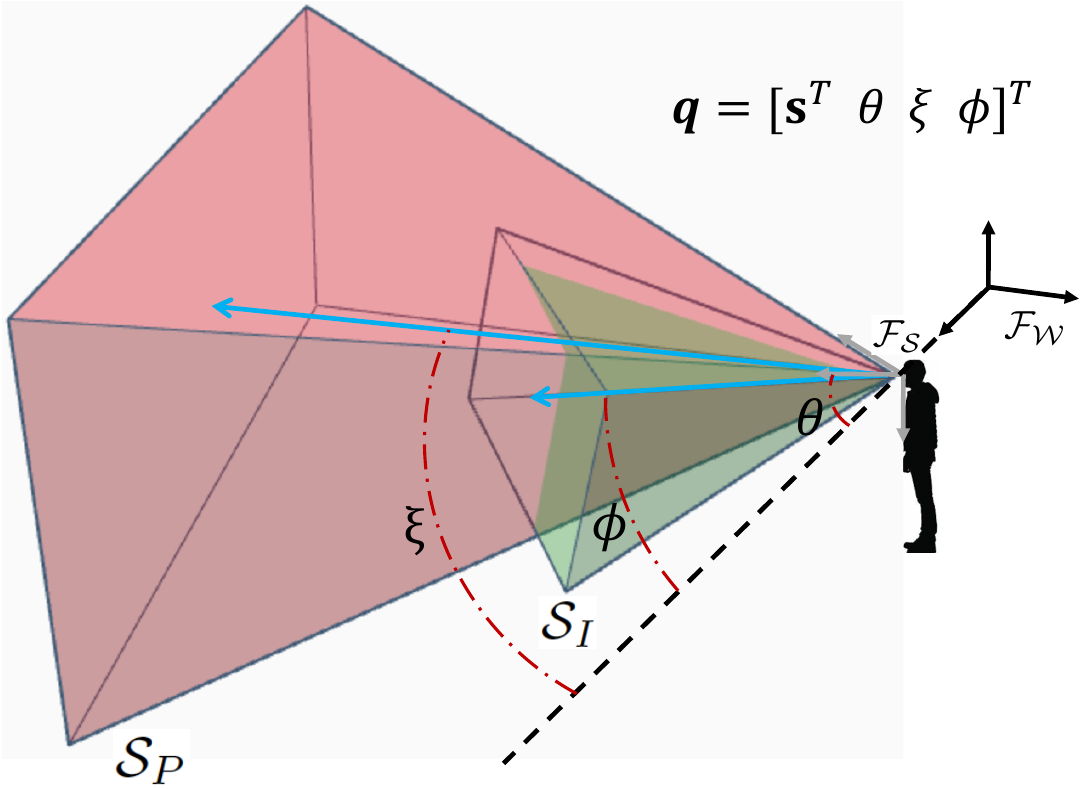}
	}
	\hfill
	\subfloat[\label{subfig:RobotFOV}]{
		\includegraphics[width = 0.45\textwidth]{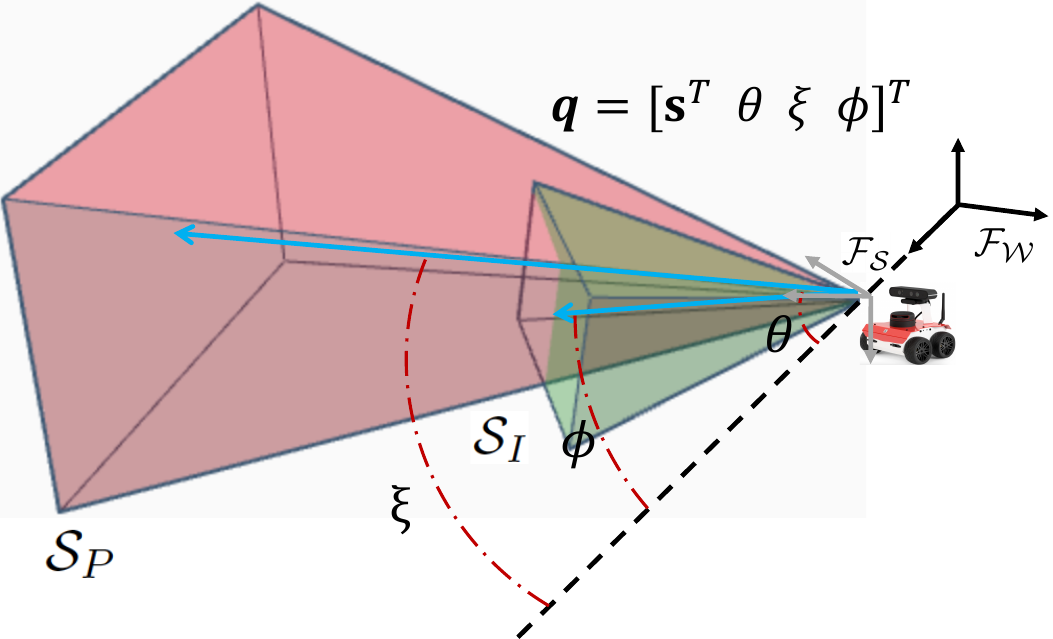}
	}
	\hfill
	\caption{Human (a) and robot (b) state, configuration, and passive and active sensor FOVs. }
	\label{fig:AgentFOVs}
\end{figure*}

Although the approach can be easily extended to other sensor configurations, in this paper it is
assumed that the information-gathering agent is equipped with one passive sensor for obstacle/target
collision avoidance and localization, with FOV denoted by $\mathcal{S}_P$, and one active sensor for
target inference and classification, with FOV denoted by $\mathcal{S}_I$ (Fig. \ref{subfig:RobotFOV}). 
In human studies, the same passive/active configuration is implemented via virtual reality (VR) 
wand/joystick and goggles, and by measuring and constraining the human FOV, as shown in 
Fig. \ref{subfig:HumanFOV}. Furthermore, the workspace is populated with $q$ known fixed, rigid, and 
opaque objects $\mathcal{B}_1,...,\mathcal{B}_q \subset \mathcal{W}$ that constitute obstacles as 
well as occlusions. Therefore, in order to observe the targets, the agent must navigate in $\mathcal{W}$ 
avoiding both collisions and occluded views, according to the following line of sight (LOS) visibility constraint:

\begin{definition}[Line of sight]
	\label{def:LineOfSight}
	Given the sensor position $\mathbf{s} \in \mathcal{W}$, a target at $\mathbf{x}\in\mathcal{W}$ is 
	occluded by an object $\mathcal{B}\subset \mathcal{W}$ if and only if,
	
	\begin{equation}\label{eqn:LineOfSight}
		L(\mathbf{s},\mathbf{x}) \cap \mathcal{B} \neq \emptyset
	\end{equation}

   \noindent
    where $L(\mathbf{s}, \mathbf{x})  = \{(1-\gamma)\mathbf{s} + \gamma \mathbf{x} ~|~ \gamma \in [0,1]\}$.
\end{definition}

Let $\mathcal{F}_{\mathcal{W}}$ denote an inertial frame embedded in $\mathcal{W}$, and 
$\mathcal{I}$ denote the geometry of the agent body . The motion of the agent relative to the 
workspace can then be described by the position and orientation of a body frame $\mathcal{F}_{\mathcal{S}}$, 
embedded in the agent, relative to $\mathcal{F}_{\mathcal{W}}$. Thus, the state of the 
information-gathering agent at $t_k$ can be described by the vector 
$\mathbf{q}_k = [\mathbf{s}_k^T \quad \theta_k \quad \xi_k\quad \phi_k]^T$, where $\mathbf{s}_k$ 
represents the inertial position of the information-gathering agent in $\mathcal{W}$, $\theta_k \in \mathbb{S}^1$ 
is the orientation of the agent, and $\xi_k\in [\xi_l,\xi_u] $ and $ \phi_k \in [\phi_l, \phi_u] $
are preferred sensing directions of the ``passive" and ``active" FOVs, respectively. In addition,
$\xi_l,\xi_u$ and $\phi_l, \phi_u$ bound the preferred sensing directions for $\mathcal{S}_P$
and $\mathcal{S}_I$ with respect to the information-gathering agent body. By this approach it is 
possible to model FOVs able to move with respect to the agent body, as required by the motion of the 
human head or pan-tilt-zoom cameras (Fig. \ref{fig:AgentFOVs}). 

Obstacle avoidance is accomplished by ensuring that the agent configuration, defined as 
$\mathbf{t}_k = [\mathbf{s}_k ^T\quad \theta_k]^T$, remains in free configuration space at all times. 
Let $\mathcal{C}$ represent all possible agent configurations, and 
$\mathcal{CB}_j = \{\mathbf{t} \in \mathcal{C} | \mathcal{I}(\mathbf{t})\cap
\mathcal{B}_j \neq \emptyset\}$ denote the C-obstacle associated with object $\mathcal{B}_j$ (defined in
\cite{FerrarInformation21} and references therein). Then, the free configuration space is the space of 
configurations that avoid collisions with the obstacles or, in other words, that are the complement of 
all C-obstacle regions in $\mathcal{C}$, i.e. $\mathcal{C}_{\text{free}} = \{\mathcal{C} \backslash \bigcup_{j=1}^q \mathcal{CB}_j\}$.

According to directional visibility theory \cite{GemerekOcclusion22}, the subset of the free space at 
which a target is visible by a sensor in the presence of occlusions can defined as follows:

\begin{definition} [Target Visibility Region]
	\label{def:TVRegion}
	For a sensor with FOV
	$\mathcal{S}_P \subset \mathcal{W}$, in the
	presence of $q$ occlusions $\mathcal{B}_j (j = 1, . . .,q)$ a target at $\mathbf{x}_i \in \mathcal{W}$ 
	is visible within the target visibility region that satisfies both FOV and LOS conditions, i.e.:
	\begin{equation}
		 \mathcal{TV}_i = \{\mathbf{t} \in \mathcal{C}_{\text{free}}~|~\mathbf{x}_i \in  \mathcal{S}_P,
		 L(\mathbf{s},\mathbf{x}_i) \cap \mathcal{B}_j = \emptyset, \forall j\}
	\end{equation}
\end{definition}

It follows that multiple targets are visible to the sensor in the intersection of multiple visibility 
regions defined as \cite{GemerekOcclusion22}:

\begin{definition} [Set Visibility Region]
	\label{def:SVRegion}
	Given a set of $r$ target-visibility regions $\{\mathcal{TV}_i ~|~ i \in \{1,2,...,r\}\}$, let $S \subseteq \{1,2,...,r\}$
	represent the set of target indices of two or more intersecting regions, such that the following holds $\bigcap_{i \in S}{\mathcal{TV}_i}
	\neq \emptyset$. Then, the set visibility region of target $i$ is defined as	
	\begin{equation}
		\mathcal{V}_S = \left\{\bigcap_{i \in S}{\mathcal{TV}_i} ~|~ S \subseteq \{1,2,...,r\}\right\}
	\end{equation}
\end{definition}

Similarly, after a target $i$ is detected and localized, the agent may observe the target features using 
the active sensor with FOV $\mathcal{S}_{I}$ provided $\mathbf{x}_i \in \mathcal{S}_{I}(\mathbf{q})$ and
$L(\mathbf{s}, \mathbf{x}_i) \cap \mathcal{B}_j = \emptyset,  \:\: 1 \leq  j \leq q $. In order to explore the 
tradeoff of information value and information cost in inferential decisions, use of the active sensor is associated 
with an information cost $J(t_k)$ that may reflect the use of processing power, data storage, and/or need for 
covertness. Then, the information-gathering agent, must make a deliberate decision to observe one or more 
target features prior to obtaining the corresponding measurement, which may consist of an image or raw 
measurement data from which feature $X_i$ may be extracted. For simplicity, measurement errors are assumed 
negligible but they may be easily introduced following the approach in \cite[Chapter~9]{FerrarInformation21}. 
Then, the goal of the treasure hunt is to infer the hypothesis variable $Y_i$ from $X_i$, $i=1,2,\ldots$, using 
a probabilistic measurement model $P(Y_i, X_{i,1}, \ldots, X_{i,n})$ \cite{FerrariInformation09}. The 
measurement model, chosen here as a Bayesian network (BN) (Fig. \ref{subfig:ActiveTargetStructure}), 
consists of a probabilistic representation of the relationship between the observed target features and the 
target classification that may be learned from expert knowledge or prior training data as shown in 
\cite[Chapter~9]{FerrarInformation21}. Importantly, because the agent may not have the time and/or 
resources to observe all target features, classification may be performed from a sequence of partial observations.

Target features are observed through test decisions made by the information-gathering agent, which
result into soft or hard evidence for the probabilistic model $P(Y_i, X_{i,1}, \ldots, X_{i,n})$ \cite{JensenBayesian07}. 
Let $u(t_k) \in \mathcal{U}_k$ denote at time $t_k$ test decision chosen from the set of all admissible tests
$\mathcal{U}_k \subset \mathcal{U}$. The set $\mathcal{U} = \{\vartheta_c,\vartheta_s,\vartheta_{un}\}$
consists of all test decisions, where $\vartheta_c$ and $\vartheta_s$ represent the decisions to continue or stop
observing target features, and $\vartheta_{un}$ represents the decision to not observe any feature. The test 
decision $u(t_{k})$ generates a measurement variable at time step $t_{k+1}$,

\begin{equation}
	z(t_{k+1}) = x_{i,l},\quad 1 \leq i \leq r, \quad 1 \leq l \leq n, \quad x_{i,l} \in \mathcal{X}_l
\end{equation}

\noindent
observed after paying the information cost $J(t_k) \in \mathbb{Z}$, which is modeled as cumulative number 
of observed features up to $t_k$. When the measurement budget  $R$ is finite, it may not be exceeded 
by the agent and, thus, the treasure hunt problem must be solved subject to the hard constraint

\begin{equation}
	J(t_k) \leq R.
\end{equation}

Action decisions modify the state of the world and/or information-gathering agent \cite{JensenBayesian07}. In
the treasure hunt problem, action decisions are control inputs that decide the position and orientation of the agent
and of the FOVs $\mathcal{S}_P$ and $\mathcal{S}_I$. Let $a(t_k) \in \mathcal{A}_k$ denote an action decision
chosen at time $t_k$ from set $\mathcal{A}_k$ of all admissible actions. The agent motion can then be described by a
causal model as the following difference equation,

\begin{equation}
    \label{eq:dynamicSys}
    \mathbf{q}_{k+1} = \mathbf{f}[ \mathbf{q}_k, a(t_k), t_k]
\end{equation}

\noindent
where $\mathbf{f}[\cdot]$ is obtained by modeling the agent dynamics.

Then, an active perception strategy consists of a sequence of action and test decisions that allow the agent to
search the workspace and obtain measurements from targets distributed therein, as follows:

\begin{definition} [Inferential Decision Strategy]
	An active inferential decision strategy is a class of admissible policies that consists of a sequence of functions,
	\begin{equation}
		\label{eq:genericstrategy}
		\sigma = \{\pi_0,\pi_1,...,\pi_T\}
	\end{equation}
\end{definition}

\noindent
where $\pi_k$ maps all past information-gathering agent states, test variables, action and test decisions
into admissible action and test decisions,

\begin{equation}
\label{eq:activetreasurehuntstrategy}
    \begin{split}
          & \{a(t_k),u(t_k)\} = \pi_k[\mathbf{q}_0,a(t_1),u(t_1),z(t_1),J(t_1),\mathbf{q}_1,\\
          & ...,a(t_{k-1}),u(t_{k-1}),z(t_{k-1}),J(t_{k-1}),\mathbf{q}_{k-1}]
    \end{split}
\end{equation}

\noindent
such that $\pi_k[\cdot] \in \{\mathcal{A}_k,\mathcal{U}_k\}$, for all $k = 1,2,...,T$.

Based on all the aforementioned definitions, the problem is formulated as follows:

\begin{problem}[Satisficing Treasure Hunt]
	\label{prob:satisficingtreasurehunt}
	
Given an initial state $\mathbf{q}_0$ and the satisificing aspiration level of total information value $\Delta$,
the satisficing treasure hunt problem consists of finding an active inferential decision making strategy, $\sigma$,
over a known and finite time horizon $(0,T]$, such that the cumulative information value collected from all 
observed features is no less than $\Delta$,

\begin{equation}
    \label{eq:problemobjectve1}
    \sum^r_{i=1} [\mathbbm{1}{(\exists k, \mathbf{x}_i \in \mathcal{S}_I(\mathbf{q}_k) \land  L(\mathbf{s}_k, \mathbf{x}_i) \cap \mathcal{B}_j, \forall  j}) I(Y_i;X_i)] \geq \Delta
\end{equation}

\noindent
where

\begin{align}
   \label{eq:motiondyn}
    & \mathbf{q}_{k+1} = \mathbf{f}[ \mathbf{q}_k, a(t_k), t_k] \\
    \label{eq:variableinfer}
    & \hat{y}_i = \argmax_{y \in \mathcal{Y}}{P(Y_i = y, X_{i,1},\ldots,X_{i,n})}\\
    \label{eq:informationvalue}
    & I(Y_i;X_i) = H(Y_i) - H(Y_i~|~X_i)\\
    \label{eq:infocostbudget}
    & J(t_T) \leq R \\
    \label{eq:targettimeindices}
    & i = 1,2,...,r, \: 1 \leq k \leq T\\
    \label{eq:obstacleindex}
    & j = 1,2,...,q
\end{align}

\end{problem}

An optimal search strategy makes use of the agent motion model (Eq. \ref{eq:motiondyn}), measurement
model (Eq. \ref{eq:variableinfer}) and knowledge of the workspace $\mathcal{W}$ to maximize the information
value while minimizing the distance traveled and the cumulative information cost \cite{FerrarInformation21}.
A feasible search strategy may use all or part of the available models of the environment and targets, or knowledge
of prior states and decisions to produce a sequence of action and test decisions that satisfy the objective
(Eq. \ref{eq:problemobjectve1}) by the desired end time $t_T$.

\section{Human Satisficing Studies}
\label{sec:Humanstudies}

Human strategies and heuristics for active perception are modeled and investigated by considering two classes
of satisficing treasure hunt problems, referred to as passive and active experiments. Passive satisficing experiments
focus on treasure hunt problems in which information is presented to the decision maker who passively observes
features needed to make inferential decisions. Active satisficing experiments allow the decision maker to control the
amount of information gathered in support of inferential decisions. Additionally, static and dynamic treasure hunt
experiments in which the agent remains stationary or navigates the workspace, respectively, are considered in order
to leverage complementary desktop and virtual reality (ambulatory) human studies.

Previous studies in cognitive psychology showed that the urgency to respond \cite{CisekDecisions09} and the need for
fast decision-making\cite{OhSatisficing16} significantly affect human decision evidence accumulation, thus leading to
the use of heuristics in solving complex problems. Passive satisficing experiments focus on test decisions, which
determine the evidence accumulation of the agent based on partial information under ``urgency". Inspired by
satisficing searches for Spanish treasures with feature ordering constraints \cite{SimonOptimal75}, active satisficing
includes both test and action decisions, which change not only the agent's knowledge and information about
the world but also its physical state. Because information gathering by a physical agent such as a human or robot is a
causal process \cite{FerrarInformation21}, feature ordering constraints are necessary in order to
describe the temporal nature of information discovery.

Both passive and active satisficing human experiments comprise a training phase and a test phase
that are also similarly applied in the robot experiments in Sections \ref{sec:Passiveperformeval}-\ref{sec:RoboApp}.
During the training phase, human participants learn the validity of target features in determining
the outcome of the hypothesis variable. They receive feedback on their inferential decisions to
aid in their learning process. During the test phase, pressures are introduced, and action decisions
are added for active tasks. Importantly, during the test phase, no performance feedback or ground truth
is provided to human participants (or robots).

\subsection{Passive Satisficing Task}
\label{subsec:Passivesatisficingexp}

\begin{figure}
    \centering
    \includegraphics[width=0.4\textwidth]{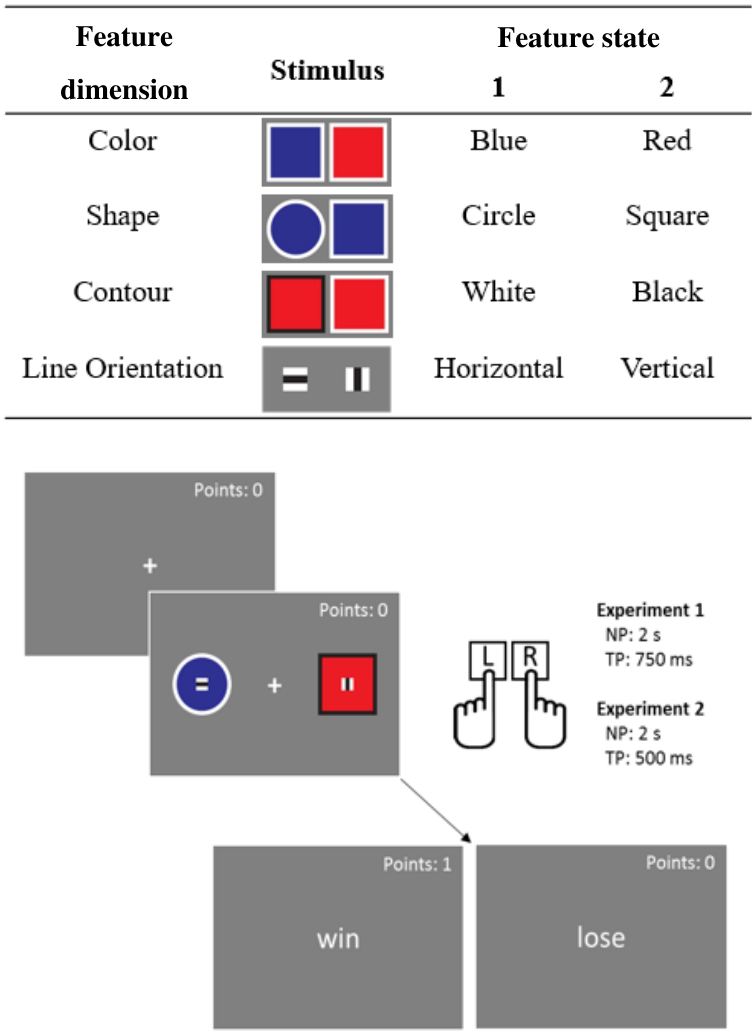}
    \caption{Features and human display used for the passive satisficing experiment, where the result of ``win" or ``lose"
    	was displayed only during the training phase.}
    \label{fig:TimePressureExp}
\end{figure}

The passive satisficing experiments presented in this paper adopted the passive treasure hunt
problem, shown in Fig. \ref{fig:TimePressureExp} and related to the well-known weather prediction
task \cite{GluckPeople02,LagnadoInsight06,SpeekenbrinkModels10}. The problem was first proposed
in \cite{OhSatisficing16} to investigate the cognitive processes involved in human test decisions under
pressure. In view of its passive nature, the experimental platform of choice consisted of a desktop computer
used to emulate the high-paced decision scenarios, and to encourage the human participants to focus on
cue(feature) combination rather than memorization \cite{OhSatisficing16,LambertsCategorization95}.

The stimuli presented on a screen were precisely controlled, ensuring consistency across participants
and minimizing distractions from irrelevant objects or external factors \cite{GarlanProject02,LavieAttention10}.
In each task, participants were presented with two different stimuli from which to select the ``treasure"
before the total time, $t_T$, at one's disposal has elapsed (time pressure). The treasures are hidden but
correlated with the visual appearance of the stimulus, and the underlying probabilities must be
learned by trial and error during the training phase. Each stimulus is characterized by four binary
cues or ``features", namely color ($X_1$), shape ($X_2$), contour ($X_3$), and line orientation
($X_4$), illustrated in the table in Fig. \ref{fig:TimePressureExp}. The goal of this passive satisficing
task is to find all treasures among stimuli that are presented on the screen or, in other words, to
infer a binary hypothesis variable $Y$, with range $\mathcal{Y} = \{y_1, y_2\}$, where $y_1 =$
``treasure" and $y_2=$ ``not treasure". The task is passive by design because the participant cannot
control the information displayed in order to aid his/her decisions.

During the training phase, each (human) participant performed 240 trials to learn the relationship between
features, $X = \{X_1, X_2, X_3, X_4\}$, and the hypothesis variable $Y$. After the training phase, participants
were divided into two groups. The first group underwent a moderate time pressure (TP) experiment and was
tested against two datasets, each consisting of 120 trials. Participants were required to make decisions within
a response time $t_T = 750$ ms, which allowed ample time to ponder on the features presented and how they related
to the treasure. The second group underwent an intense TP experiment, with a response time of only $t_T = 500$
ms. Participants in this group also encountered two datasets, each containing 120 trials. A more detailed
description of the experiment, including redundant features and human subject procedures, can be found in \cite{OhSatisficing16}.
Subsequently, the task was modified to develop a number of active satisficing treasure hunts in which information
about the treasures had to be obtained by navigating a complex environment, as explained in the next section.

\subsection{Active Satisficing Treasure Hunt Task}
\label{subsec:Activesatisficingexp}

\begin{figure}
    \centering
      \subfloat[\label{subfig:ActiveTrainingNoReveal}]{
      	\includegraphics[width = 0.23\textwidth]{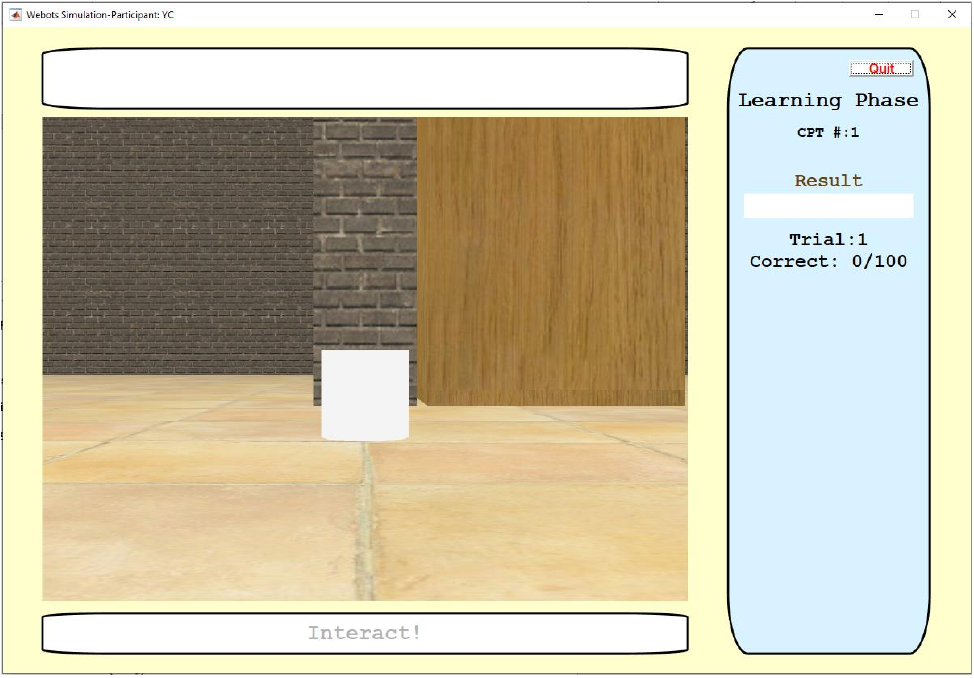}
      }
      \hfill
      \subfloat[\label{subfig:ActiveTrainingFeatureReveal}]{
      	\includegraphics[width = 0.23\textwidth]{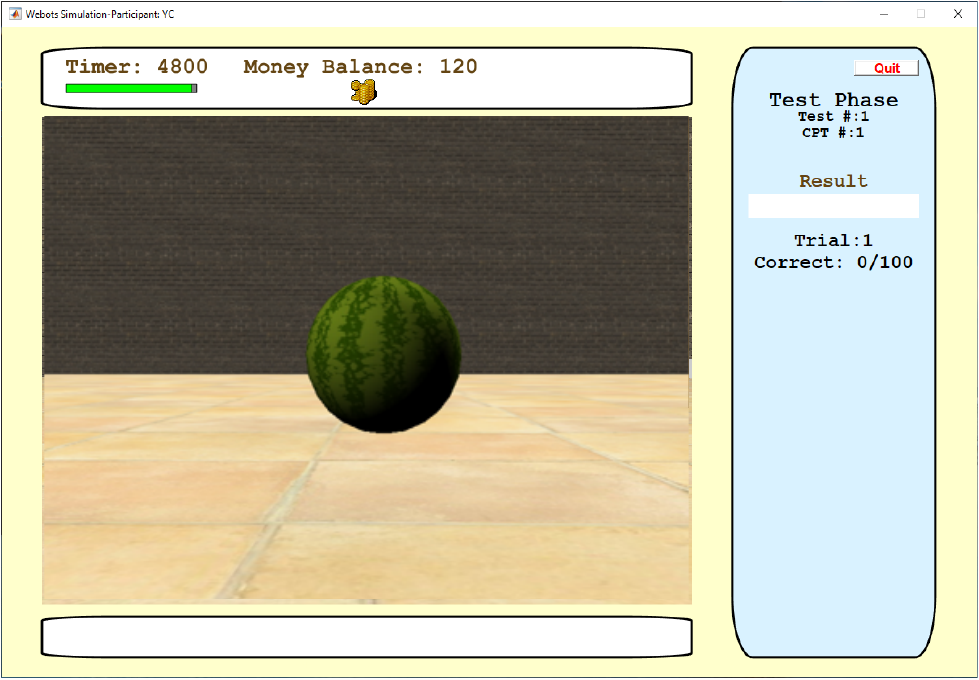}
      }
    \hfill
        \subfloat[\label{subfig:ActiveWorkspace}]{
    	\includegraphics[width = 0.2\textwidth]{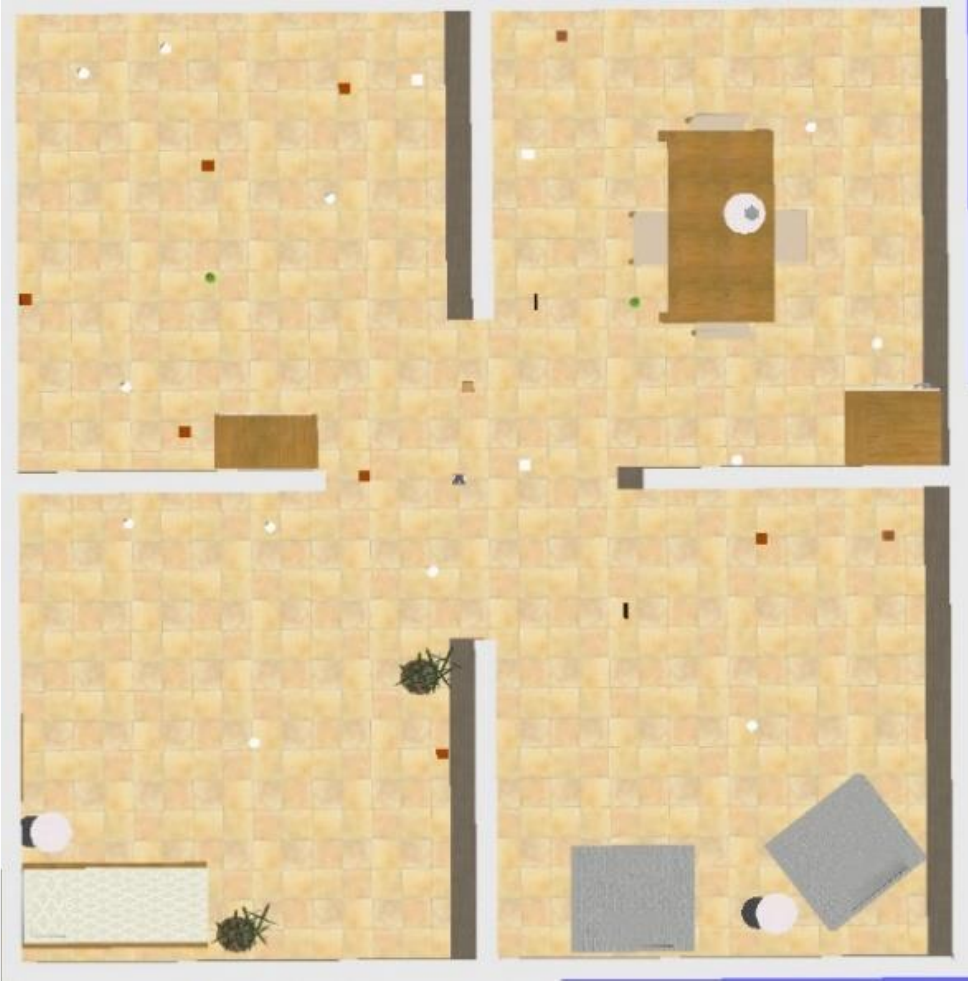}
    }
    \hfill
        \subfloat[\label{subfig:ActiveTargetStructure}]{
    	\includegraphics[width = 0.26\textwidth]{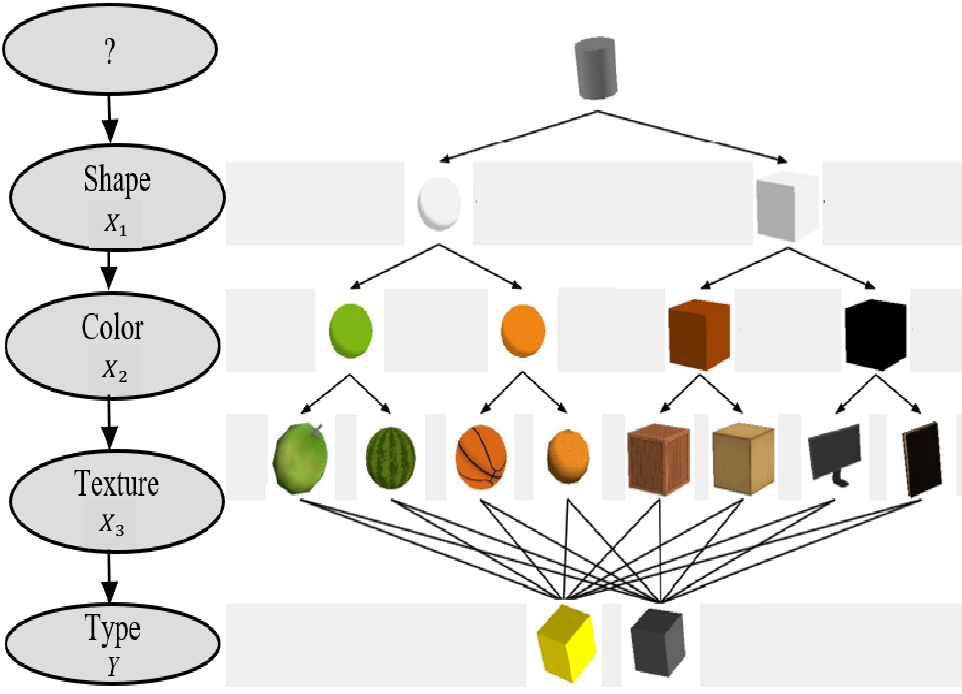}
    }
    \hfill
    \caption{First-person view in training phase without prior target feature revealed (a) and
     with feature revealed by a participant (b) in the Webots\textregistered$\,$
    workspace (c) and target features encoded in a BN structure with ordering constraints (d).}
    \label{fig:ActiveSatisficingExp}
\end{figure}

The satisficing treasure hunt task is an ambulatory study in which participants must navigate a complex
environment populated with a number of obstacles and objects in order to first find a set of targets (stimuli) and,
then, determine which are the treasures. Additionally, once the targets are inside the participant's FOV, features are
displayed sequentially to him/her only after paying cost for the information requested. The ordering constraints
(illustrated in Fig. \ref{subfig:ActiveTargetStructure}) allow for the study of information cost and its role in the
decision making process by which the task is to be performed not only under time pressure but also a fixed budget.
Thus, the satisficing treasure hunt allows not only to investigate how information about a hidden variable (treasure)
is leveraged, but also how humans mediate between multiple objectives such as obstacle avoidance, limited sensing
resources, and time constraints. Participants must, therefore, search and locate the treasures without any prior
information on initial target features, target positions, or workspace and obstacle layout.

In order to utilize a controlled environment that can be easily changed to study all combinations of features,
target/obstacle distributions, and underlying probabilities, the active satisficing treasure hunt task was developed
and conducted in a virtual reality environment known as the DiVE \cite{DukeDiVE22}. By this approach different
experiments were designed and easily modified so as to investigate different difficulty levels and provide the
human participants repeatable, well-controlled, and immersive experience of acquiring and processing
information to generate behavior \cite{VanNavigating98,PanWhyandhow18,ServotteVirtual20}.
The DiVE consists of a 3m x 3m x 3m stereoscopic rear projected room with head and hand tracking,
allowing participants to interact with a virtual environment in real-time \cite{DukeDiVE22}. By developing
a new interface between the DiVE and the robotic software Webots$^\circledR$, this research was able to
readily introduce humans within the same environments designed for humans, and vice versa, according
to the BN model of the desired treasure hunt task. The structure of the BN used for the human/robot treasure 
hunt perception task is plotted in Fig. \ref{subfig:ActiveTargetStructure}. The BN parameters, not shown for 
brevity, were varied across trials to obtain a representative dataset from the human study from which 
mathematical models of human decision strategies could be learned and validated.

Six human participants were trained and given access to the DiVE for a total of fifty-four trials
with the objective to model aspects of human intelligence that outperform existing robot strategies.
The number of trials and participants is adequate to the scope of the study which was not to learn
from a representative sample of the human population, but to extract inferential decision making
strategies generalizable to treasure hunt robot problems. Besides manageable in view of the high
costs and logistical challenges associated with running DiVE experiments, the size of the resulting
dataset was also found to be adequate to varying all of the workspace and target characteristics
across experiments, similarly to the studies in \cite{ZiebartMaximum08,LevineNonlinear11}.
Moreover, through the VR googles and environment, it was possible to have precise and controllable
ground truth not only about the workspace, but also about the human FOV, $\mathcal{S}_P$, within
which the human could observe critical information such as targets, features, and obstacles.

A mental model of the relationship between target features and classification was first learned by
the human participants during 100 stationary training sessions (Fig. \ref{subfig:ActiveTrainingNoReveal}
and Fig. \ref{subfig:ActiveTrainingFeatureReveal}) in which the target features (visual cues), comprised of
shape ($X_1$), color ($X_2$), and texture ($X_3$), followed by the target classification  $Y$,
where $\mathcal{Y} = \{y_1,y_2\}$, were displayed on a computer screen, through the desktop Webots$^\circledR$
simulation shown in Fig. \ref{fig:ActiveSatisficingExp}. Participants were then instructed to search for
treasures inside an unknown 10m x 10m Webots$^\circledR$ workspace with $r=30$ targets
(Fig. \ref{subfig:ActiveWorkspace}), by paying information cost $J(t_k)$ to see the features,
$X_i = \{X_{i,1}, X_{i,2}, X_{i,3}\}$, of every target (labeled by $i$)  inside their FOV sequentially
over time (test phase). Based on the features observed, which may have included one or more features
in the set $X$, participants were asked to decide which targets were treasures ($Y = y_1$) or not
($Y = y_2$). No feedback about their decisions was provided and, as explained in
Section \ref{sec:Probformulation}, the task had to be performed within a limited budget $R$
and time period $t_T$.

\begin{figure}
	\centering
	\subfloat[\label{subfig:ActiveTesting2}]{
		\includegraphics[width = 0.4\textwidth]{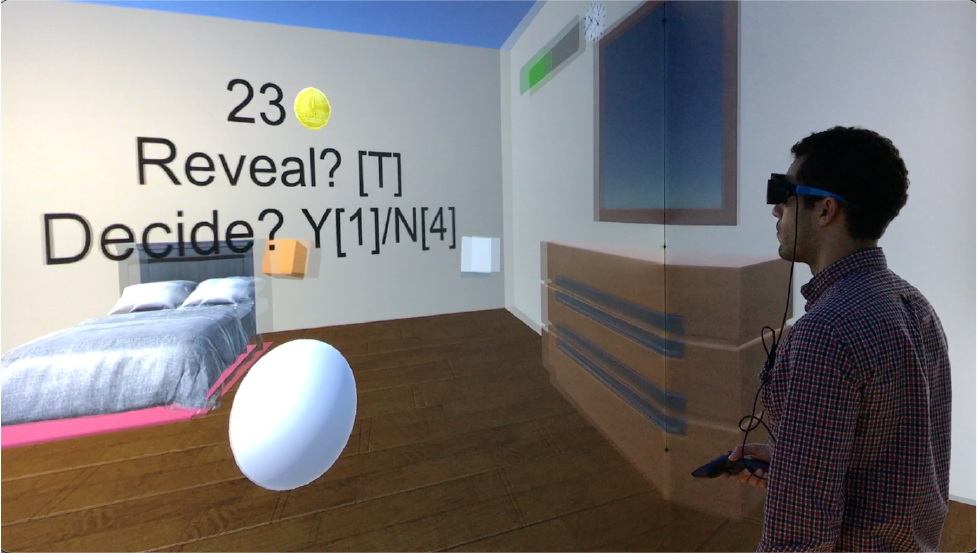}
	}
	\hfill
	\subfloat[\label{subfig:ActiveTesting1}]{
		\includegraphics[width = 0.23\textwidth]{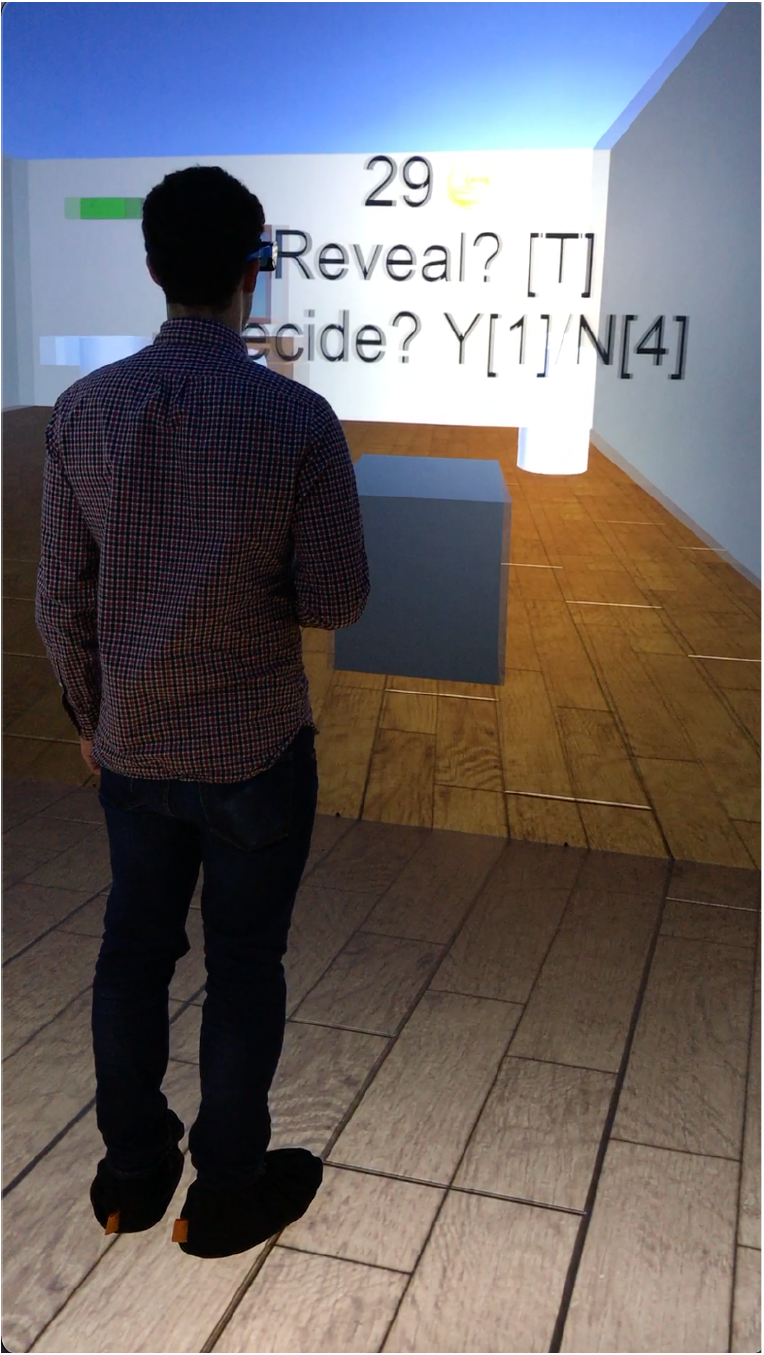}
	}
	\hfill
	\caption{Test phase in active satisficing experiment in DiVE. }
	\label{fig:ActiveTesting}
\end{figure}

Mobility and ordering feature constraints are both critical to autonomous sensors and robots,
because they are intrinsic to how these cyber-physical systems gather information and interact with the world
around them. Thanks to the simulation environments and human experiment design presented in
this section, we were able to engage participants in a series of classification tasks in which target
features were revealed only after paying both a monetary and time cost, similarly to artificial sensors
that require both computing and time resources to process visual data. Participants were able to build a
mental model built for decision making with the inclusion of temporal constraints during the training
phase, according to the BN conditional probabilities (parameters) of each study. By sampling the
Webots$^\circledR$ environments from each BN model, selected by the experiment designer to encompass
the full range of inference problem difficulty, and by transferring them automatically into VR
(Fig. \ref{fig:ActiveTesting}) the data collected was guaranteed ideally suited for the modeling and
generalization of human strategies to robots (Section \ref{sec:Activestratgymodeling}).
As explained in the next section, the test phase was conducted under three conditions:
no pressure, money pressure, and sensory deprivation (fog).

\section{External Pressures Inducing Satisficing}
\label{sec:Externpressures}

Previous work on human satisficing strategies and heuristics illustrated that most humans resort to
these approaches for two main reasons, one is computational feasibility and the other is the ``less-can-be-more"
effect \cite{GigerenzerHeuristic11}. When the search for information and computation costs
become impractical for making a truly ``rational" decision, satisficing strategies adaptively drop
information sources or partially explore decision tree branches, thus accommodating the limitations of
computational capacity. In situations in which models have significant deviations from the ground truth,
external uncertainties are substantial, or closed-form mathematical descriptions are lacking, optimization
on potentially inaccurate models can be risky. As a result, satisficing strategies and heuristics often
outperform classical models by utilizing less information. This effect can be explained in two ways. Firstly,
the success of heuristics is often dependent on the environment. For example, empirical evidence suggests
that strategies such as ``take-the-best," which rely on a single good reason, perform better than classical
approaches under high uncertainty \cite{HogarthHeuristic07}. Secondly, decision-making systems should
consider trade-offs between bias and variance, which is determined by model complexity\cite{BishopPattern06}.
Simple heuristics with fewer free parameters have smaller variance than complex statistical models,
thus avoiding overfitting to noisy or unrepresentative data, and generalizable across a wider range of datasets \cite{BishopPattern06,BrightonBayesian08,GigerenzerHomo09}.

Motivated by the situations where robots' mission goals can be severely hindered or completely
compromised due to inaccurate environment or sensing models caused by pressures, the paper
seeks to emulate aspects of human intelligence under the pressures and study their influence on
decisions. The environment pressures include, for example, time pressure \cite{PayneAdaptive88},
information cost \cite{DieckmannInfluence07, BroderDecision03}, cue(feature) redundancy
\cite{DieckmannInfluence07, RieskampSelectStrategy06}, sensory deprivation, and high risks
\cite{SlovicAffect05, PorcelliStress17}. Cue(feature) redundancy and high risk have been investigated
extensively in statistics and economics, particularly in the context of inferential decisions
\cite{KruschkeBayesian10, MullainathanBehavioral00}. In the treasure hunt problem, sensory
deprivation and information cost directly and indirectly influence action decisions, which brings
insight how these pressures impact agents' motion. However, the effects of sensory deprivation on
human decisions have not been thoroughly investigated compared to other pressures.  Time pressure
is ubiquitous in the real world, yet heuristic strategies derived from human behavior are still lacking.
Thus, this paper aims to fill this research gap by examining the time pressure, information cost pressure,
and sensory deprivation and their effects on decision outcomes.

\subsection{Time Pressure}
\label{subsec:Timepressure}

Assume that a fixed time interval $t_c$ is needed to integrate one additional feature into the inference
decision-making process. In the meantime, each decision must be made within $t_T$, and $\mathit{p}_i$
is the number of observed features for the $i$th target. The satisficing strategies must adaptively select
a subset of the features such that a decision is made within the time constraint

\begin{equation}
    \label{eq:timepressure}
    \mathit{p}_i t_c < t_T, i = 1,2,...,r
\end{equation}

According to the human studies in \cite{OhSatisficing16}, the response time of participants in the
passive satisficing tasks was measured during the pilot work. The average response time in these tasks
was found to be approximately 700 ms. Based on this finding, three time windows were designed to
represent different time pressure levels: a two-second time window was considered without any time
pressure; a 750 ms time window was considered moderate time pressure; and a 500 ms time window
was considered intense time pressure.

\subsection{Information Cost}
\label{subsec:Infocost}

The cost of acquiring new information intrinsically makes an agent use fewer features to reach a decision.
In Section \ref{sec:Probformulation}, new information for the $i$th target is collected through a sequence
of $\mathit{p}_i$ observed target features. Thus, for all $r$ targets, the information cost is
mathematically described as the total number of observed features not exceeding a preset
budget $R$

\begin{equation}
    \label{eq:informationcost}
    \sum_{i=1}^{r}\mathit{p}_i \leq R
\end{equation}

In Section \ref{subsec:Activesatisficingexp}, the human studies introduce information cost pressure
using the parameter $R = 30$. In the context of the treasure hunt problem, $R$ represents the
measurement budget, which limits the number of features that a participant can observe from targets.
In this experiment, for example, a total of $r = 30$ targets was used,  and an information budget of $R = 30$
was chosen such that the human participants were able to observe, on average, one feature per target.
Other experiments were similarly performed by considering a range of parameters that spanned task
difficulty levels across participants and treasure hunt types.

\subsection{Sensory Deprivation}
\label{subsec:Sensorydeprivation}

\begin{figure}[t]
    \centering
    \subfloat[\label{subfig:SimFogEnv}]{
        \includegraphics[width=0.38\textwidth]{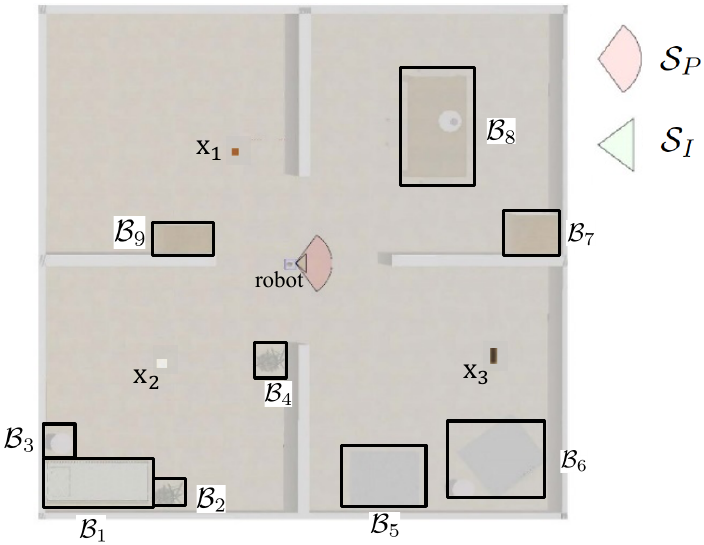}
    }
    \hfill
    \subfloat[\label{subfig:SimFogFPV}]{
        \includegraphics[width=0.32\textwidth]{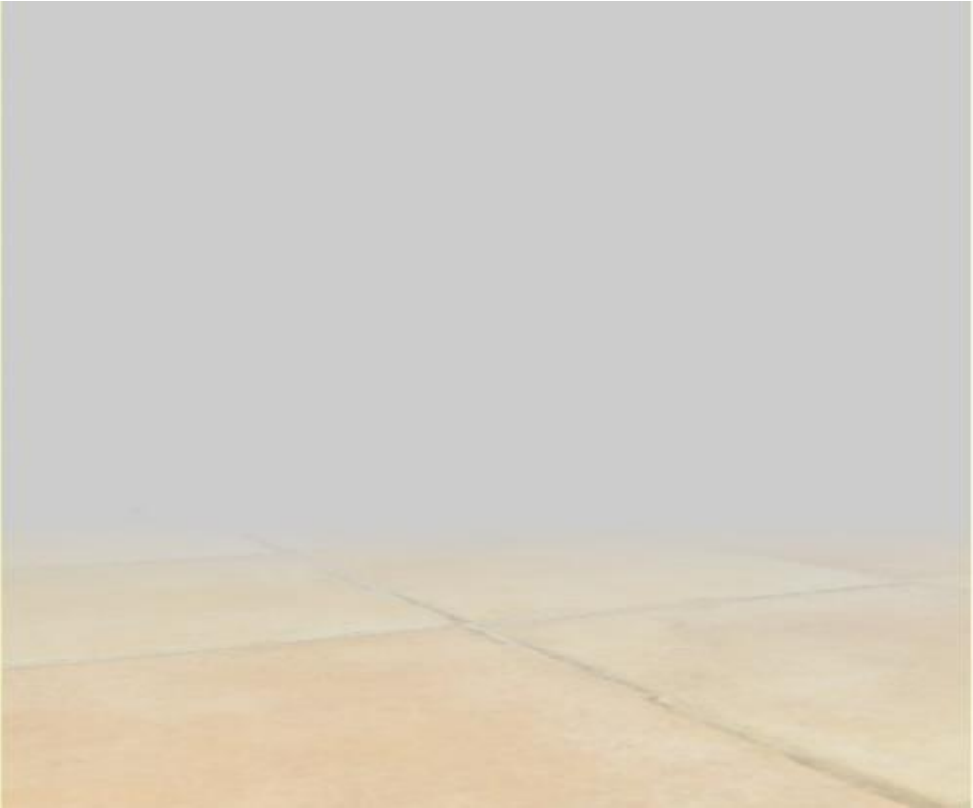}
    }
    \caption{Top view visibility conditions of the unknown workspace (a) and first-person view of poor visibility
    	condition (b) due to fog.}
    \label{fig:SimFogEnv}
\end{figure}

As explained in Section \ref{sec:Probformulation}, information-gathering agents were not provided a map
of the workspace $\mathcal{W}$ \emph{a priori}, and, instead, were required to obtain information about
target and obstacle positions and geometries by means of a passive on-board sensor (e.g. camera or LIDAR)
with FOV $\mathcal{S}_P$ as shown in Fig. \ref{subfig:SimFogEnv}. From the definition of set visibility
region (Definition \ref{def:SVRegion}), for a subset $S \subseteq \{1,2,...,r\}$ of target indices, the set  visibility
region $\mathcal{V}_{S} \subseteq \mathcal{C}_{\text{free}}$ contains all targets in $S$ visible to passive
sensor with FOV $\mathcal{S}_P$. A globally optimal solution to treasure hunt problem
(Eq. \ref{eq:problemobjectve1}-\ref{eq:obstacleindex}) with respect to a subset of targets $S$ is feasible if
and only if $\mathcal{V}_S \neq \emptyset$.

In parallel to the human studies in Section \ref{subsec:Activesatisficingexp},
robot sensory deprivation was introduced by  simulating/producing fog in the workspace, thereby reducing the
FOV radius to approximately 1m, in a 20m x 20m robot workspace. A fog environment is simulated inside the
Webots\textregistered$ $ environment as shown in Fig. \ref{fig:SimFogEnv}, thereby reducing the camera's ability
to view targets inside the sensor $\mathcal{S}_P$. As a result, $V_{S} = \emptyset$ even when there are $|S|=2$ targets,
indicating that a globally optimal solution is infeasible. Consequently, optimal strategies typically fail under
sensory deprivation due to lack of target information. Using the methods presented in the next section, human
strategies for modulating between satisficing and optimizing strategies are first learned from data and, then,
generalized to autonomous robots, as shown in Section \ref{sec:RoboApp}. Satisficing strategies are aimed at
overcoming this difficulty, and use local information to explore the environment and visit targets.

\begin{figure*}[t]
    \centering
    \subfloat[\label{subfig:TimePressureHumStudy1}]{
    \includegraphics[width = 0.7\textwidth]{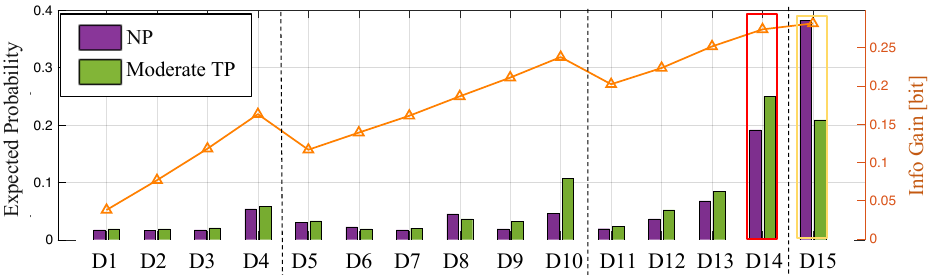}
    }
    \hfill
    \subfloat[\label{subfig:TimePressureHumStudy2}]{
    \includegraphics[width = 0.7\textwidth]{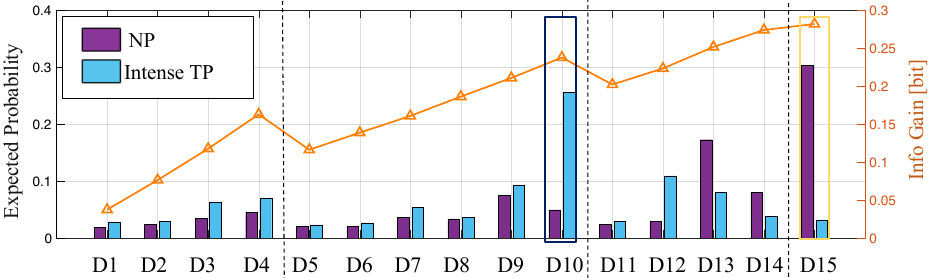}
    }
     \hfill
    \subfloat[\label{subfig:TimePressureHumStudy3}]{
    \includegraphics[width = 0.7\textwidth]{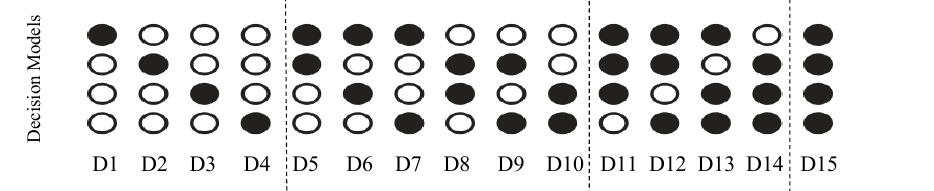}
    }
    \caption{Human data analysis results for  the moderate TP experiment (a) and
     the intense TP experiment (b) with the enumeration of decision models (c).}
    \label{fig:TPInfo}
\end{figure*}

\section{Mathematical Modeling of Human Passive Satisficing Strategies}
\label{sec:Passivestrategymodeling}

Previous work by the authors showed that human participants drop less
informative features to meet pressing time deadlines that do not allow them to complete the tasks optimally \cite{OhSatisficing16}.
The analysis of data obtained from the moderate TP experiment (Fig. \ref{subfig:TimePressureHumStudy1})
and intense TP experiment (Fig. \ref{subfig:TimePressureHumStudy2}) reveals similar interesting
findings regarding human decision-making under different time pressure conditions. Under the no TP
condition, the most probable decision model selected by human participants (indicated by yellow boxes in Fig.
\ref{subfig:TimePressureHumStudy1} and Fig. \ref{subfig:TimePressureHumStudy2}) utilizes all four features
and aims at maximizing information value. However, under moderate TP, the most probable decision model
selected by human participants (indicated by a red box in Fig. \ref{subfig:TimePressureHumStudy1}) uses
only three features and has lower information value than the no TP condition. As time pressure becomes the most stringent in
the intense TP, the most probable decision model selected by human participants (indicated by a dark blue box in
Fig. \ref{subfig:TimePressureHumStudy2}) uses only two features and exhibits even lower information value
than observed in the previous two time pressure conditions. These results demonstrate the trade-offs made
by human participants among time pressure, model complexity, and information value. As time pressure increases,
individuals adaptively opt for simpler decision models with fewer features, and sacrificed information value to meet
the decision deadline, thus reflecting the cognitive adaptation of human participants in response to time constraints.

\subsection{Passive Satisficing Decision Heuristic Propositions}
\label{subsec:Passivemodelpropsition}

Inspired by human participants' satisficing behavior indicated by the data analysis above, this paper develops
three heuristic decision models, which accommodate varying levels of time pressure and adaptively select a
subset of information-significant features to solve the inferential decision making problems. For simplicity and 
based on experimental evidence, it was assumed that observed features were error free.

\subsubsection{Discounted Cumulative Probability Gain (ProbGain)}
\label{subsubsec:Modelprobgain}

The heuristic is designed to incorporate two aspects of behaviors observed from human data. First, the
heuristic encourages the use of features that provide high information value for decision-making. By
summing up the information value of each feature, the heuristic prioritizes the features that contribute
the most to evidence accumulation. Second, the heuristic also considers the cost of using multiple features
in terms of processing time. By applying a higher discount to models with more features, the heuristic
discourages excessive cost on time that might lead to violation of time constraints.

For an inferential decision-making problem with \textbf{sorted} $\mathit{p}$ observed features $\{x_j\}_{j=1}^p$
according to the information value $v_I(x_j)$ in descending order, where $v_I(x_j)$ representing the increase
in information value with respect to the maximum a-posterior rule

\begin{equation}
        \label{eq:probabilitygain}
        v_I(x_j) =  \max_{y \in \mathcal{Y}}p(Y = y~|~x_j) - \max_{y \in \mathcal{Y}}p(Y = y)
\end{equation}

Let $\{x_{1},x_{2},...,x_{i}\}$ represent a subset of observed features that contains the first ($i$)
most informative features with respect to  $v_I(x_j)$, where $t_T$ is the allowable time to make a classification
decision, and the discount factor $\gamma(t_T) \in (0,1)$ is defined to be a function of $t_T$ to represent the
penalty induced by time pressure. Then, the heuristic strategy can be modeled as follows,

\begin{equation}
    \label{eq:heuristic1}
    H_{\text{ProbGain}}(t_T,\{x_j\}_{j=1}^\mathit{p}) = \argmax_{i} \{\gamma(t_T)^{i}\sum_{j = 1}^{i} v_I(x_{j})\}
\end{equation}

\noindent
where

\begin{equation}
    \label{eq:TimeDiscount}
    \gamma(t_T) = \exp(-\frac{\lambda}{t_T})
\end{equation}

\subsubsection{Discounted Log-odds Ratio (LogOdds)}
\label{subsubsec:Modellogodds}

Log odds ratio plays a central role in classical algorithms like logistic regression \cite{BishopPattern06},
and represents the ``confidence" of making a inferential decision. The update of log odds ratio with respect to
a ``new feature" is through direct summation, thus taking advantage of the feature independence and arriving at
fast evidence accumulation. Furthermore, the use of log odds ratio in the context of time pressure is
slightly modified such that a discount is applied with inclusion of an additional feature to penalize the feature
usage because of time pressure. By combining the benefits of direct summation for fast evidence accumulation
and the discount for time pressure as inspired from human behavior, the heuristic based on log odds ratio can
make efficient decisions by considering the most relevant features under time constraints.

For an inferential decision-making problem with \textbf{sorted} $\mathit{p}$ observed features $\{x_j\}_{j=1}^\mathit{p}$
according to the information value $~|~v_I(x_j)~|~$ in descending order, where $~|~v_I(x_j)~|~$ represents the
log odds ratio of observed features $x_j$.  Then, the heuristic strategy can be modeled as follows,

\begin{equation}
	\label{eq:heuristic2}
	H_{\text{LogOdds}}(t_T,\{x_j\}_{j=1}^\mathit{p}) = \argmax_{i}\{\gamma(t_T)^i~|~v_0 + \sum_{j=1}^i v_I(x_j)~|~\}
\end{equation}

\noindent
where

\begin{equation*}
    v_I(x_j) = log(p(x_j~|~y_1)) - log(p(x_j~|~y_2))
\end{equation*}

\begin{equation*}
    \label{eq:H2InitBias}
    v_0 = log(p(Y=y_1)) - log(p(Y=y_2))
\end{equation*}

\subsubsection{Information Free Feature Number Discounting (InfoFree)}
\label{subsubsec:Modelinfofree}

The previous two feature selection heuristics are both based on comparison: multiple candidate sets of features are
evaluated and compared, and the heuristics select the one with the best trade-off between information value
and processing time cost. A simpler heuristic is proposed to avoid comparisons and reduces the computation burden,
while still showing the behavior that dropping less informative features due to time pressure observed from human
participants.

Sort the $\mathit{p}$ features according to the information value $v_I(x_j)$ in descending order as 
$x_{1},x_{2},...,x_{\mathit{p}}$, and a subset of  the first $i$ most informative features refers to as 
$\{x_{1},x_{2},...,x_{i}\}$. The heuristic strategy is as follows

\begin{equation}
         \label{eq:heuristic3}
        H_{\text{InfoFree}}(t_T) = \ceil*{\mathit{p} \exp(-\frac{\lambda}{t_T})}
\end{equation}

The outputs of the three heuristics are the numbers of features to be fed into the model $P(Y_i, X_{i,1}, \ldots, X_{i,n})$ 
to make an inference decision. Some mathematical properties (e.g., convergence and
monotonicity) of the three proposed heuristic strategies are presented in Appendix \ref{app:TPMathproperties}.

\subsection{Model Fit Test Against Human Data}
\label{subsec:Passivedatafit}

\begin{figure}
    \centering
    \includegraphics[width=0.45\textwidth]{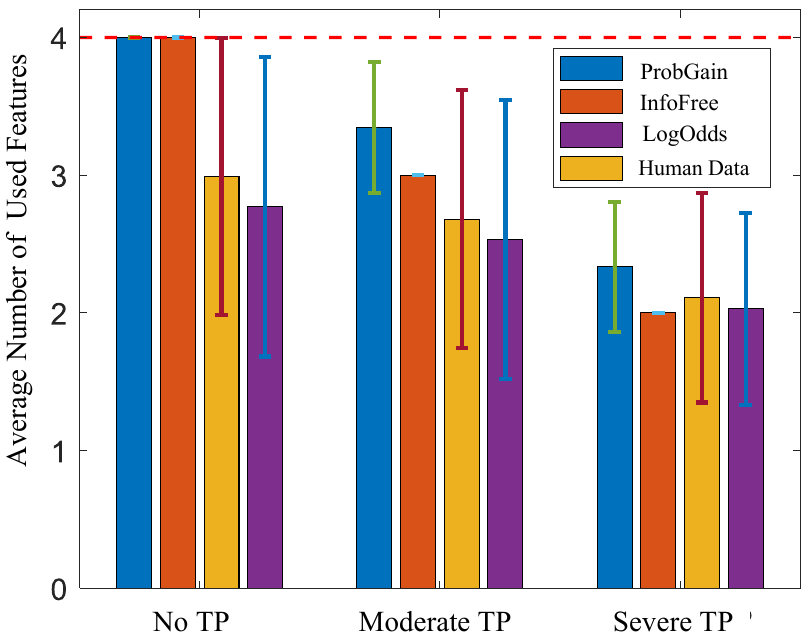}
    \caption{Mean and standard deviation of the number of used features of three heuristic
    strategies and the human strategy under three time pressure levels.}
    \label{fig:TimePressureModeling}
\end{figure}

The model fit tests against human data of the three proposed time-adaptive heuristics are under
three time pressure levels, with the time constraints scaled to ensure comparability between human
experiments and heuristic tests. The results, as shown in Fig. \ref{fig:TimePressureModeling}, indicate
two major observations. First, as time pressure increases, all three strategies utilize fewer features, thus
demonstrating their adaptability to time constraints and mirroring the behavior observed in human participants.
Second, among the three strategies, $H_{\text{LogOdds}}$ exhibits the closest average number of features and
standard deviation to the human data across all time pressure conditions. Consequently, $H_{\text{LogOdds}}$
is the heuristic strategy that best matches the human data among the three proposed strategies.

\section{Autonomous Robot Applications of Passive Satisficing Strategies}
\label{sec:Passiveperformeval}

The effectiveness of the human passive satisficing strategies modeled in the previous section, namely the three 
heuristics denoted by $H_{\text{ProbGain}}$, $H_{\text{LogOdds}}$, and $H_{\text{InfoFree}}$, was tested 
on an autonomous robot making inferential decisions on the well-established database known as car evaluation 
dataset \cite{DuaCarEval2019}. This dataset, containing 1728 samples, is chosen over other benchmark problems 
because its size is comparable to the database used for modeling human heuristics and is characterized by six 
possibly redundant features, which allows for the ability to adaptively select a subset of features to infer the 
target class. The performance of the three heuristics is compared against that of a naïve Bayes classifier, referred 
to as ``Bayes optimal" herein, which utilizes all available features for decision-making.

The car evaluation dataset records the cars' acceptability, on the basis of six features and originally
 four classes. The four classes are merged into two. A training set of 1228 samples is used to learn
 the conditional probability tables (CPTs), ensuring equal priors for both classes. After learning the CPTs,
 500 samples are used to test the classification performance of the heuristics and the naïve Bayes classifier.
 The tests are conducted under three conditions: no TP, moderate TP, and intense TP.

\begin{figure}
    \centering
    \includegraphics[width=0.44\textwidth]{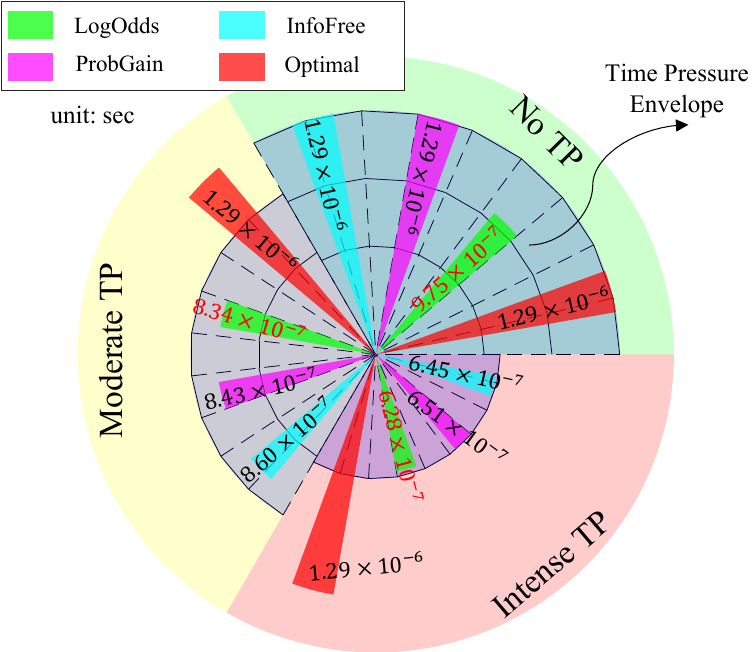}
    \caption{Processing time (unit: sec) of three time-adaptive
    heuristics and the ``Bayes optimal" strategy.}
    \label{fig:TimeNeededTP}
\end{figure}

The experiments are performed on a digital computer using MATLAB R2019b on an AMD Ryzen 9
3900X processor. The processing times of the strategies are depicted in Fig. \ref{fig:TimeNeededTP}.
If a heuristic's processing time falls within the time pressure envelope (blue area), the time
constraints are considered satisfied. The no TP condition provides sufficient time for all heuristics
to utilize all features for decision-making. The moderate TP condition allows for 75\% of the time available
in the no TP condition, whereas the intense TP condition allows for 50\% of the time available in the
no TP condition. All three heuristics are observed to satisfy the time constraints across all time pressure conditions.

\begin{figure}
    \centering
    \includegraphics[width=0.44\textwidth]{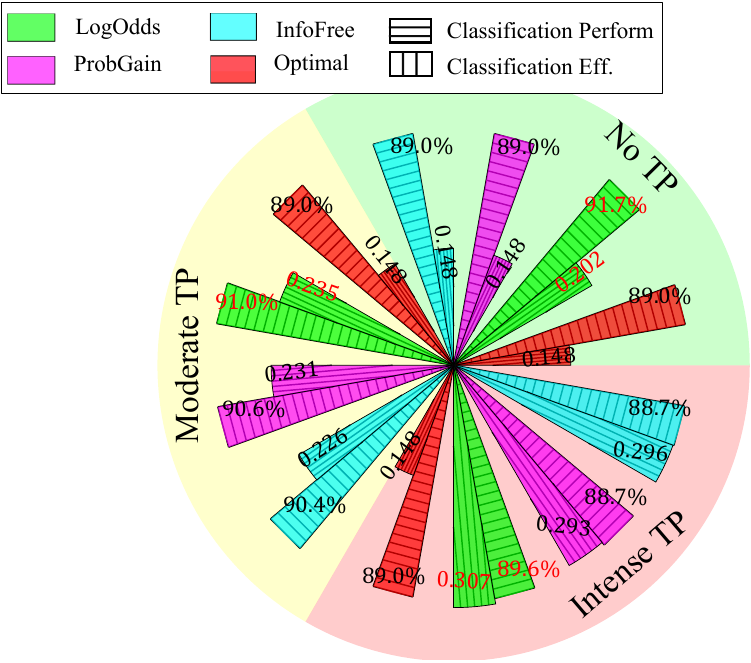}
    \caption{Classification performance and efficiency of three
    time-adaptive heuristics under three time
    pressure conditions. }
    \label{fig:TPClassificationPerform}
\end{figure}

The classification performance and efficiency of the
three time-adaptive strategies is plotted in Fig. \ref{fig:TPClassificationPerform}. $H_{\text{LogOdds}}$ outperforms 
the other three strategies on this dataset, and its performance deteriorates as time pressure increases. Under 
moderate TP, the three time-adaptive strategies use fewer features but achieve better classification performance 
than Bayes optimal. This finding exemplifies the less-can-be-more effect \cite{GigerenzerHeuristic11}. The classification
efficiency measures the average contribution of each feature to the classification performance. Bayes optimal
displays the lowest efficiency, because it utilizes all features for all time pressure conditions, whereas
$H_{\text{LogOdds}}$ exhibits the highest efficiency among the three heuristics across all time
pressure conditions.

\section{Mathematical Modeling of Human Active Satisficing Strategies}
\label{sec:Activestratgymodeling}

In the active satisficing experiments, human participants face pressures due to (unmodelled)
information cost (money) and sensory deprivation (fog pressure). These pressures prevent the
participants from performing the test and action decisions optimally. The data analysis results
for the information cost pressure, as described in Section \ref{subsec:Moneypressuremodeling},
reveal that the test decisions and action decisions are coupled. The pressure on test decisions
affect the action decisions made by the participants. The data analysis of the sensory deprivation
(fog pressure) does not incorporate existing decision-making models, such as
\cite{ZiebartMaximum08,LevineNonlinear11,GhahramaniLearningDBN06,PutermanMarkov90},
because the human participants perceive very limited information, thus violating the assumptions
underlying these models. Instead, a set of decision rules are extracted in the form of heuristics
from the human participants data from inspection. These heuristics capture the
decision-making strategies used by the participants under sensory deprivation (fog pressure).

\subsection{Information Cost (Money) Pressure}
\label{subsec:Moneypressuremodeling}

Previous studies showed that, when information cost was present, humans used a single good reason
strategy (e.g., take-the-best) in larger proportion than compensatory strategies, which integrated
all available features, to make decisions \cite{DieckmannInfluence07}; and information cost induced
humans to optimize decision criteria and shift strategies to save cost on inferior features\cite{BroderDecision03}.
This section analyzes the characteristics of human decision behavior under information cost pressure
compared with no pressure condition.

Two separate assumptions are made regarding the incentives that influence human decision-making
under information cost(money) pressure. First, based on the previous studies \cite{CaiCellDecomposition09, FerrarInformation21,
	ZhangInformation09} on ``treasure hunt" problem, the action and test decisions are driven by three
separate objectives:  the information value ($B$), the information cost ($J$), and the human travel
distance cost ($D$). Thus, it is assumed that human participants aim to maximize an objective function
(Eq. \ref{eq:treasurehuntobj}), which is a weighted sum of the three objectives.  The corresponding
weights $\omega_B$, $\omega_D$, and $\omega_J$ represent the human participants' assigned
importance to each objective

\begin{equation}
    \label{eq:treasurehuntobj}
    V = \sum_{k=0}^{T} \omega_B B(t_k) - \omega_D D(t_k) - \omega_J J(t_k)
\end{equation}

\begin{figure}[b]
     \centering
    \subfloat[\label{subfig:MoneyPressureOptAss1}]{
        \includegraphics[width=0.23\textwidth]{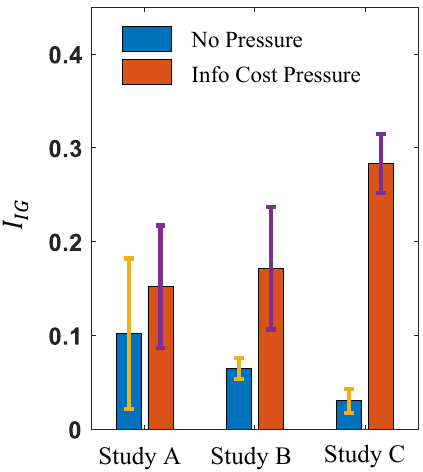}
    }
    \hfill
    \subfloat[\label{subfig:MoneyPressureOptAss2}]{
        \includegraphics[width=0.23\textwidth]{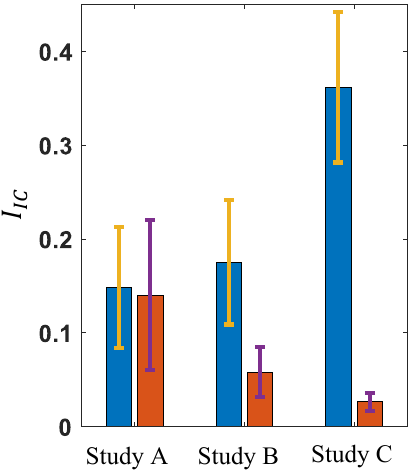}
    }
    \caption{(a) Information value attempt index $I_{IG}$ and (b) information cost
    parsimony index $I_{IC}$ of high performance human participants under two pressure conditions.}
    \label{fig:MoneyPressureIRL}
\end{figure}

The averaged weights utilized by human participants are estimated using the Maximum Entropy
Inverse Reinforcement Learning algorithm, adopted from \cite{ZiebartMaximum08}, in order
to understand the effects of money pressure on human decision behaviors. The two indices,
$I_{IG} = \omega_B / \omega_D$ and $I_{IC}  = \omega_B / \omega_J$, are designed by calculating
the ratios of the three averaged weights, and they reflect the incentives underlying human decision
behaviors. $I_{IG}$, the information value attempt index, measures the willingness of human
participants to trade travel distance for information value. $I_{IC}$, the information cost parsimony
index, measures the willingness of human participants to spend ``money" (i.e., incur costs) for
information value. The results of the analysis (shown in Fig. \ref{fig:MoneyPressureIRL})
indicate that under the information cost (money) pressure condition, human participants are
more willing to travel longer distances to acquire information value (higher $I_{IG}$). However,
they are less willing to incur costs (lower $I_{IC}$) for information value, thus suggesting a
tendency to be more conservative in spending resources for information acquisition.

\begin{figure}
    \centering
    \includegraphics[width=0.23\textwidth]{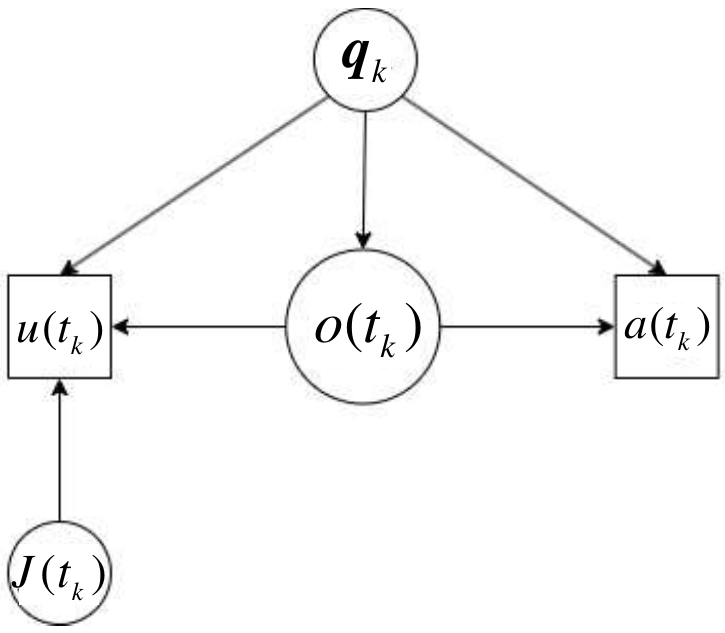}
    \caption{The intra-slice DBN that models human decision behavior}
    \label{fig:DBNIntra}
\end{figure}

\begin{figure}[b]
    \centering
    \includegraphics[width=0.4\textwidth]{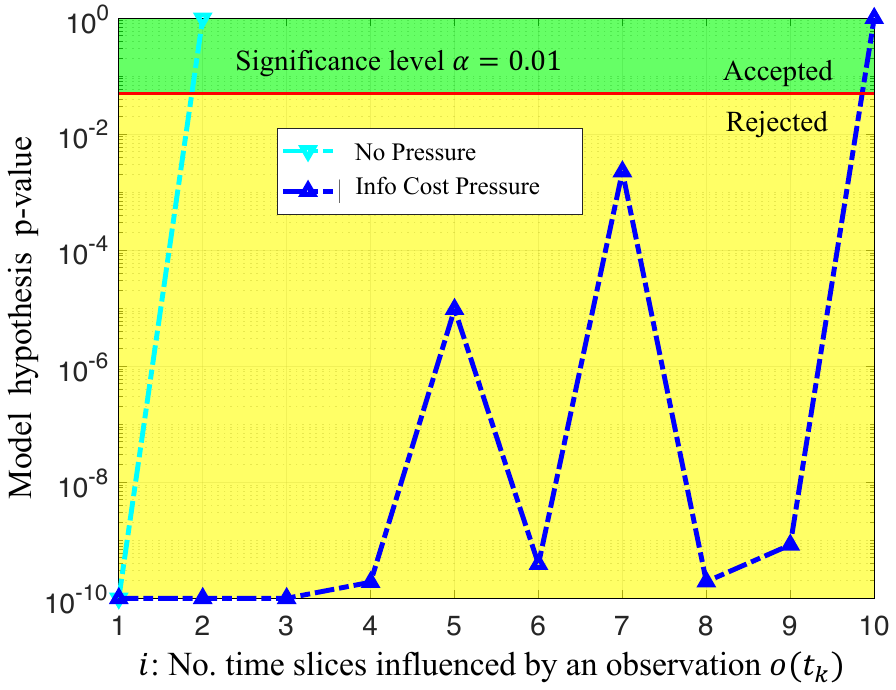}
    \caption{DBN inter-slice structure hypothesis testing results}
    \label{fig:MoneyPressureDBNAss}
\end{figure}

Furthermore, it is assumed that no utility is associated with states or actions in human decision-making.
Instead, decisions are made on this basis of causal relationships. To model human decision behavior under
this assumption, this paper uses dynamic Bayesian networks (DBNs). The DBN intra-slice structure, as
shown in Fig. \ref{fig:DBNIntra}, includes variables such as the human participants' states $\mathbf{q}_k$,
the action decision $a(t_k)$, the test decision $u(t_k)$, the set of visible targets $o(t_k)$ at time $t_k$,
and the ``money"(information cost) already spent $J(t_k)$. The intra-slice variables capture the relevant information for
decision-making at a specific time slice. This paper investigates the inter-slice structure to understand
how observations influence subsequent action and test decisions. The key question is: in how many time
slices does an observation $o(t_k)$ influence decision-making? To determine the appropriate inter-slice
structure, this paper conducts a series of hypothesis tests to assess the conformity of various models
against the human decision data. Fig. \ref{fig:MoneyPressureDBNAss} presents the results of these
hypothesis tests. Each data point represents a $p$-value that evaluates the null hypothesis:
``model $i+1$ does not fit the human data significantly better than model $i$". The models are defined
according to the number of time slices in which an observation influences decisions. If the $p$-value is
smaller than the significance level $\alpha$, the null hypothesis is rejected, thus indicating that the
subsequent model fits the data better than the previous one.

According to the results plotted in Fig. \ref{fig:MoneyPressureDBNAss}, under the no pressure condition,
an observation $o(t_k)$ influences one subsequent decision. However, under the information cost(money) 
pressure, an observation $o(t_k)$ influences nine subsequent decisions. This finding suggests that the influence of
observations extends over a longer time horizon under information cost(money) pressure than in the no pressure condition.

\subsection{Sensory Deprivation (Fog Pressure)}
\label{subsec:Fogpressuremodeling}

The introduction of sensory deprivation (fog pressure) in the environment poses two main difficulties for
human participants during navigation. First, fog limits the visibility range, thus hindering human participants'
capability of locating targets and being aware of obstacles. Second, fog impairs spatial awareness, thus
hindering human participants' ability to accurately perceive their own position within the workspace.

\begin{figure}
    \centering
    \subfloat[\label{subfig:HumanFogBehave1}]{
    \includegraphics[width = 0.48\textwidth]{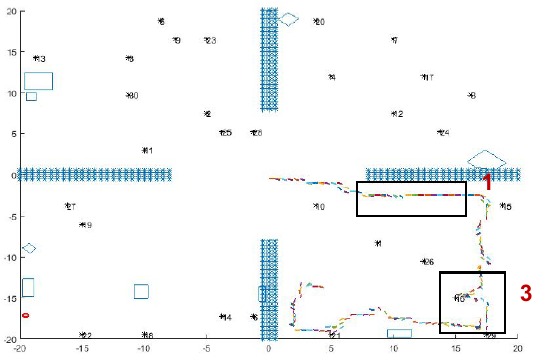}
    }
    \hfill
    \subfloat[\label{subfig:HumanFogBehave2}]{
    \includegraphics[width = 0.45\textwidth]{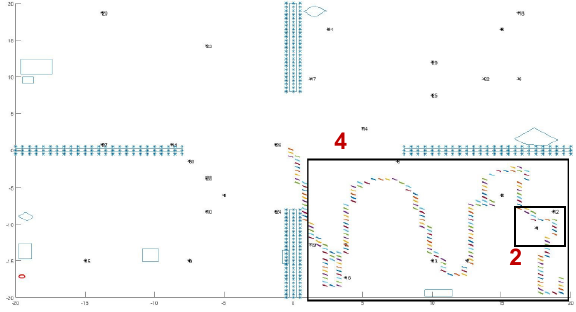}
    }
    \hfill
    \subfloat[\label{subfig:HumanFogBehave3}]{
    \includegraphics[width = 0.45\textwidth]{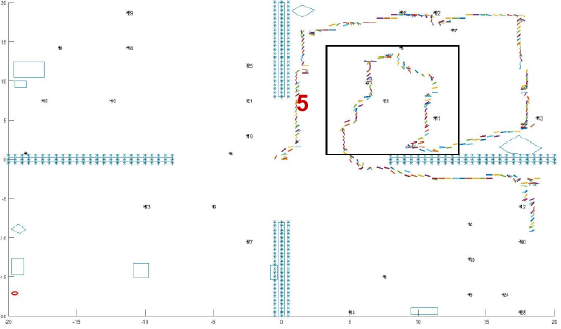}
    }
    \hfill
    \subfloat[\label{subfig:HumanFogBehave4}]{
    \includegraphics[width = 0.45\textwidth]{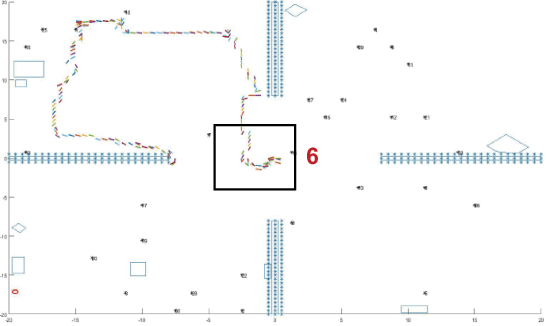}
    }
    \hfill
    \caption{The human behavior patterns in a fog environment.}
    \label{fig:HumanFogBehave}
\end{figure}

In situations in which target and obstacle information is scarcely accessible and uncertainties are 
difficult to model, human participants were found to use local information to navigate the workspace, 
observe features, and classify all targets in their FOVs \cite{GigerenzerHeuristic11, DieckmannInfluence07}.
By analyzing the human decision data collected through the active satisficing experiments described in 
Section \ref{sec:Humanstudies}, significant behavioral patterns shared by the top human performers 
can be summarized by the following six behavioral patterns exemplified by the sample studies 
plotted in Fig. \ref{fig:HumanFogBehave}:

\begin{enumerate}
	
    \item When participants enter an area and no targets are immediately visible, they follow the 
    walls or obstacles detected in the workspace (Fig. \ref{subfig:HumanFogBehave1}).

    \item When participants detect multiple targets, they pursue targets one by one, prioritizing them 
    by proximity (Fig. \ref{subfig:HumanFogBehave2}).

    \item While following a wall or obstacle, if participants detect a target, they will deviate
    from their original path and pursue the target, and may then return to their previous
    ``wall/obstacle follow" path after performing classification (Fig. \ref{subfig:HumanFogBehave1}).

    \item Upon entering an enclosed area (e.g. room), participants may engage in a strategy of covering 
    the entire room (Fig. \ref{subfig:HumanFogBehave2}).

    \item After walking along a wall or obstacle for some time without encountering any targets, participants 
    are likely to switch to a different exploratory strategy (Fig. \ref{subfig:HumanFogBehave3}).

    \item In the absence of any visible targets, participants may exhibit random walking behavior (Fig. \ref{subfig:HumanFogBehave4}).

\end{enumerate}

Detailed analysis of the above behavioral patterns (omitted for brevity) showed that the following 
three underlying incentives drive human participants in the presence of fog pressure:

\begin{itemize}
	
    \item \textbf{Frugal}: Human participants exhibit tendencies to avoid repeated visitations.
    Navigating along walls or obstacles helps participants localize themselves by using walls or obstacles as reference points.

    \item \textbf{Greedy}: Human participants demonstrate a strong motivation to find targets and engage
    with them. After a target is detected, participants pursue it and interact with it immediately.

    \item \textbf{Adaptive}: Human participants display adaptability by using multiple strategies for exploring
    the workspace. These strategies include ``wall/obstacle following,"  ``area coverage," and ``random walk."
    Participants can switch among these strategies according to the effectiveness of their current approach in finding targets.

\end{itemize}

Based on these findings, a new algorithm referred to as \textbf{AdaptiveSwitch}
(Algorithm \ref{alg:AdaptiveSwitch}) was developed to emulate humans' ability to transition 
between the three heuristics when sensory deprivation prevents the implementation of optimizing 
strategies. The three exploratory heuristics consist of wall/obstacle following ($\pi_1$), area coverage ($\pi_2$),
and random walk ($\pi_3$). The probability of executing each heuristic is referred to as $\Pi = [b_1, b_2, b_3]^T$,
where $b_i$ represents the probability of executing $\pi_i$. The index $g$ indicates the exploratory policy
being executed, and $k$ represents the number of steps taken while executing a policy. The maximum number
of steps before updating the distribution $\Pi$ is $K$. The policy for interacting with targets is
$\pi_I(u(t_k)~|~\mathbf{q}_k, o(t_k))$, and the policy for pursuing a target is $\pi_P(a(t_k)~|~\mathbf{q}_k, o(t_k))$.

\begin{algorithm}
\caption{AdaptiveSwitch}\label{alg:AdaptiveSwitch}
\begin{algorithmic}[1]
\STATE $\Pi = [b_1,b_2,b_3]^T$
\STATE $k=0, g = 0$
\WHILE{($t_k \leq t_T$ $\lor$  not all targets are classified)}
\IF{$\exists \mathbf{x}_j \in \mathcal{S}_I(\mathbf{q}_k)$}
\STATE $\pi_I(u(t_k)~|~\mathbf{q}_k, o(t_k))$
\ELSE
\IF{$o(t_k) \neq \emptyset$}
\STATE $\pi_P(a(t_k)~|~\mathbf{q}_k, o(t_k))$
\STATE $k = 0, g = 0$
\ELSE
\IF {$g > 0$\,  $\land$ \,$k \leq K$}
\STATE $\pi_{g}(a(t_k)~|~\mathbf{q}_k, o(t_k))$
\STATE $k = k + 1$
\ELSE
\IF {$k \geq K$}
\STATE $\Pi[g] = \gamma*\Pi[g]$
\ELSE
\IF {not closed to wall}
\STATE $\Pi[1] = 0$
\ELSE
\STATE $\Pi[1] = \beta(b_1 + b_2 + b_3)$
\ENDIF
\ENDIF
\STATE $\Pi = $normalize$(\Pi)$
\STATE $g \sim \Pi$
\ENDIF
\ENDIF
\ENDIF
\ENDWHILE
\end{algorithmic}
\end{algorithm}

As shown in Algorithm \ref{alg:AdaptiveSwitch}, the greediness of the heuristic strategy (lines 4 - 9) captures 
the behaviors in which participants interact with targets if possible (line 4) and pursue a target if it is visible (line 7). 
If no targets are visible and the maximum exploratory step $K$ is not exceeded, the current exploratory heuristic 
continues to be executed (lines 11-13). The adaptiveness of the three exploratory heuristics is shown in lines 15 - 22. 
If the current exploratory heuristic is executed for more than $K$ steps, its probability of execution is discounted 
(line 16). The probability of executing the ``wall/obstacle following" heuristic increases if the participant is close 
to a wall/obstacle; otherwise this heuristic is disabled (lines 19 - 21). 

\begin{figure}
	
    \centering
    \includegraphics[width=0.5\textwidth]{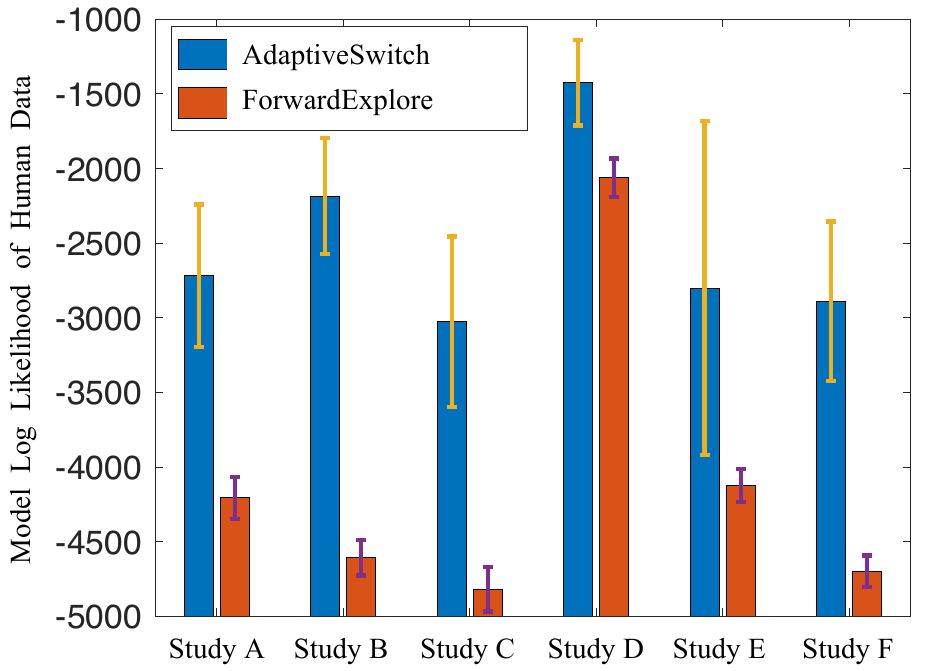}
    \caption{Averaged model log likelihood of AdaptiveSwitch and ForwardExplore
    	in six human studies.}
    \label{fig:FogHeuristicLoglik}

\end{figure}

After learning the parameters from the human data, the \textbf{AdaptiveSwitch} algorithm was compared to 
another hypothesized switching logic referred to as \textbf{ForwardExplore} in which participants
predominantly move forward with a high probability and turn with a small probability or when encountering an 
obstacle. In order to determine which switching logic best captured human behaviors, the log likelihood of 
AdaptiveSwitch and ForwardExplore was computed using the human data from the active satisficing experiment 
involving six participants. The results plotted in Fig. \ref{fig:FogHeuristicLoglik} show that the log likelihood of 
AdaptiveSwitch is greater than that of ForwardExplore across all human experiment trials. This finding suggests 
that AdaptiveSwitch aligns more closely with the observed human strategies than ForwardExplore and, therefore, 
was implemented in the robot studies described in the next section.

\section{Autonomous Robot Applications of Active Satisficing Strategies}
\label{sec:RoboApp}

Two key contributions of this paper are the applications of the modeled human strategies on a robot, and the comparison
of optimal strategies and the modeled human strategies in pressure conditions, under which
optimization is infeasible. For simplicity, the preferred sensing directions of $\mathcal{S}_P$ and $\mathcal{S}_I$
are assumed to be fixed with respect to the robot platform. Therefore, the state vector for a robot reduces to
$\mathbf{q} = [x \quad y \quad \theta]^T$, where the orientation of the robot platform $\theta$ also represents the
preferred sensing directions. Both sensor FOVs are modeled by sectors with angle-of-view
$\zeta_1,\zeta_2 \in [0,2\pi)$ and radii $r_1, r_2 > 0$. The two FOVs share the same apex and their bisectors
coincide with each other.

\subsection{Information Cost (Money) Pressure}
\label{subsec:Infcostapp}
\begin{figure*}
	\centering
	\subfloat[\label{subfig:MoneyPressureStudy1}]{
		\includegraphics[width = 0.3\textwidth]{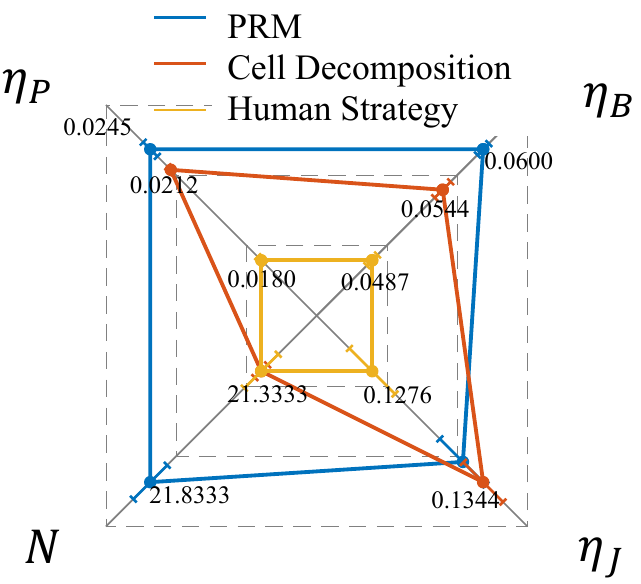}
	}
	\hfill
	\subfloat[\label{subfig:MoneyPressureStudy2}]{
		\includegraphics[width = 0.3\textwidth]{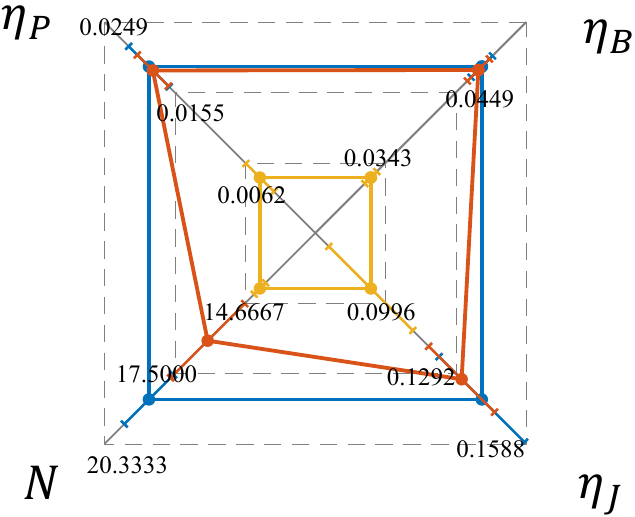}
	}
	\hfill
	\subfloat[\label{subfig:MoneyPressureStudy3}]{
		\includegraphics[width = 0.3\textwidth]{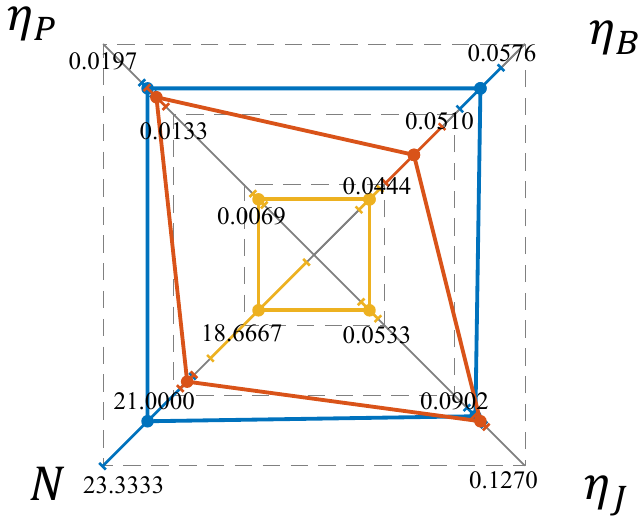}
	}
	\hfill
	\subfloat[\label{subfig:MoneyPressureStudy4}]{
		\includegraphics[width = 0.3\textwidth]{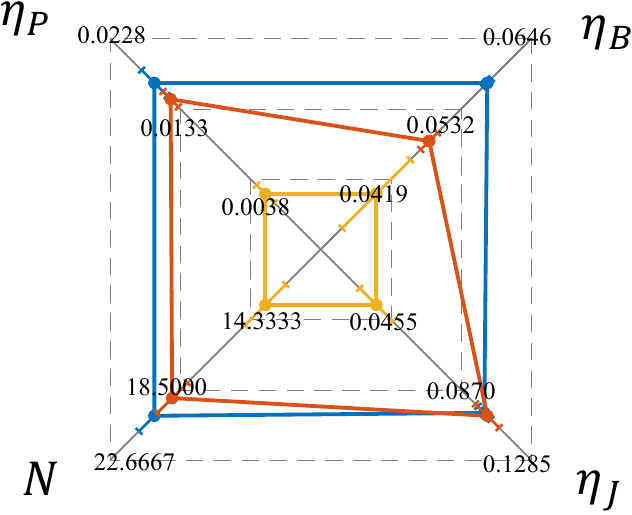}
	}
	\hfill
	\subfloat[\label{subfig:MoneyPressureStudy5}]{
		\includegraphics[width = 0.3\textwidth]{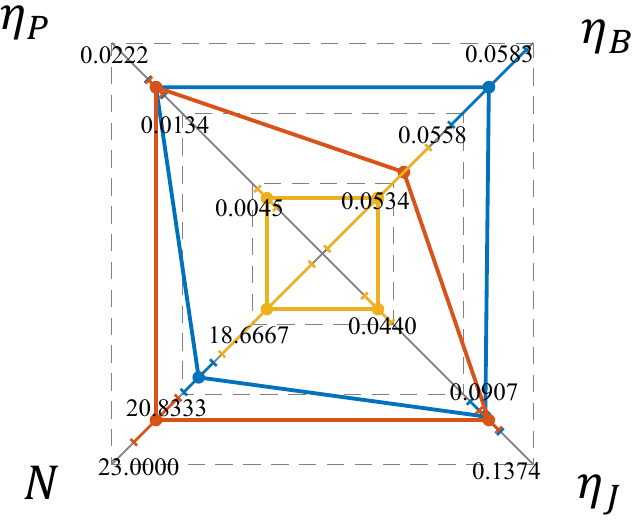}
	}
	\hfill
	\subfloat[\label{subfig:MoneyPressureStudy6}]{
		\includegraphics[width = 0.3\textwidth]{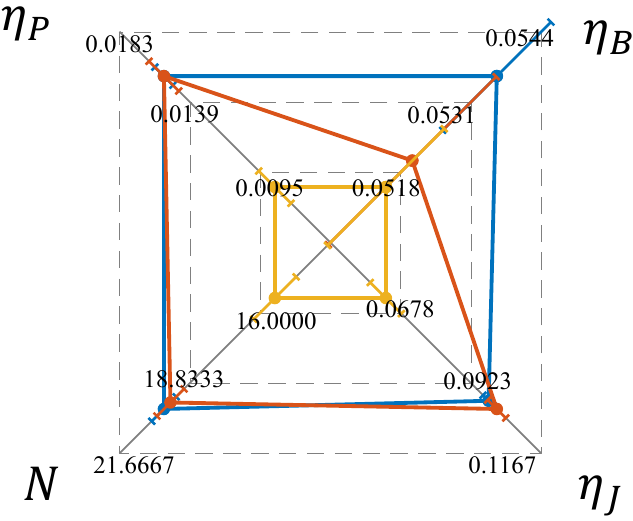}
	}
	\caption{Performance comparison of two optimal strategies and human strategy
		over six case studies (a)-(f).}
	\label{fig:MoneyPressureStudies}
\end{figure*}

The introduction of information cost increases the complexity of planning test decisions.
In the absence of information cost, a greedy policy that observes all available features for
any target is considered ``optimal",  because it collects all information value without any
cost. However, when information cost is taken into account, a longer planning horizon for
test decisions becomes crucial to effectively allocate the budget for observing features
of all targets. This paper proposes two planners, a probabilistic roadmap (PRM) based
planner and a cell decomposition based planner, to solve the robot treasure hunt problem, which has
identical workspace, initial conditions, and target layouts to the problem faced by human participants
in the active satisficing experiment. The objective function Eq. \ref{eq:treasurehuntobj} is
maximized by using these methods. Unlike existing approaches \cite{CaiCellDecomposition09, FerrariInformation09, ZhangInformation09}
that solve the original version of the treasure hunt problem as described in \cite{FerrarInformation21},
the developed planners handle the problem without pre-specification of the final robot configuration.
Consequently, the search space increases exponentially, thus rendering label-correcting algorithms
\cite{BertsekasDynamicProgramming12} no longer applicable. Additionally, unlike previous methods
that solely optimize the objective with respect to the path, the developed planners consider the constraint
 on the number of observed features due to information cost pressure.
The number of observed features thus becomes a decision variable with a long planning horizon.
To solve the problem, the developed planners use PRM and cell decomposition techniques to generate
graphs representing the workspace \cite{FerrarInformation21}. The Dijkstra algorithm is used to
compute the shortest path between targets. Furthermore, an MINLP algorithm is used to determine the
optimal number of observed features and the visitation sequence of the targets.

\subsubsection{Performance Comparison with Human Strategies}
\label{subsubsec:Moneypressureperfom}

The performance (Fig. \ref{fig:MoneyPressureStudies}) of two optimal strategies
(PRM and cell decomposition) is compared with the human strategy from human
data. Under information cost (money) pressure, the path and the number of
observed features for each target are optimized with respect to a linear combination
of three objectives. Defining$\tau$ as a planned continuous path
\cite{LavallePlanningAlgorithmsBook06}, this paper focuses on four
performance metrics: path efficiency $\eta_{P} = 1/D(\tau)$ [$m^{-1}$];
information gathering efficiency $\eta_{B} = B(\tau)/D(\tau)$ [$\text{bit}/m$];
measurement productivity $\eta_{J} = B(\tau)/J(\tau)$ [bit]; and classification
performance $N = N(\tau)$. Higher numbers are better for all metrics. Six case studies
are examined. One case study comprises of three different experiment layouts. The
optimal strategies and the human participants have no prior knowledge of the target
positions and initial features, and all environmental information is obtained from
FOV $\mathcal{S}_P$. The results, shown in Fig. \ref{fig:MoneyPressureStudies},
indicate that the two optimal strategies consistently outperform the human strategy
across all four performance metrics. The performance envelopes of the optimal strategies
are outside of the performance of the human strategy, thus indicating their superiority.

The finding that the optimal strategies outperform human strategies is unsurprising,
because information cost (money) pressure imposes a constraint on only the expenditure
of measurement resources, which can be effectively modeled mathematically. The finding
suggests that under information cost (money) pressure, near-optimal strategies can make
better decisions than human strategies.

\subsection{Sensory Deprivation (Fog Pressure)}
\label{subsec:Fogpressureapp}

An extensive series of tests are conducted to evaluate the effectiveness of AdaptiveSwitch
(Section \ref{sec:Activestratgymodeling}) under sensory deprivation(fog) conditions and compare it with other
strategies. These tests comprise of 118 simulations and physical experiments, encompassing
various levels of uncertainty. The challenges posed by fog in robot planning are twofold. First,
fog obstructs the robot's ability to detect targets and obstacles by using onboard sensors such
as cameras, thus making long-horizon optimization-based planning nearly impossible. Second,
fog complicates the task of self-localization for the robot with respect to the entire map, although
short-term localization can rely on inertial measurement units. Three test groups are described
as follows:

\subsubsection{Performance Tests in the Human Experiment Workspace}
\label{subsubsec:Fogtestshumanws}

AdaptiveSwitch is applied to the workspace and target layouts in the active satisficing experiment
workspace described in Fig. \ref{fig:ActiveTesting} and Fig. \ref{fig:SimFogEnv}. The
experiment involves six human participants, each of whom completes three trials with different target
layouts, thus resulting in a total of eighteen different target layouts with a uniform obstacle layout.

\begin{figure}[t]
	
	\centering
	\subfloat[\label{subfig:FogHumanNv}]{
		\includegraphics[width = 0.45\textwidth]{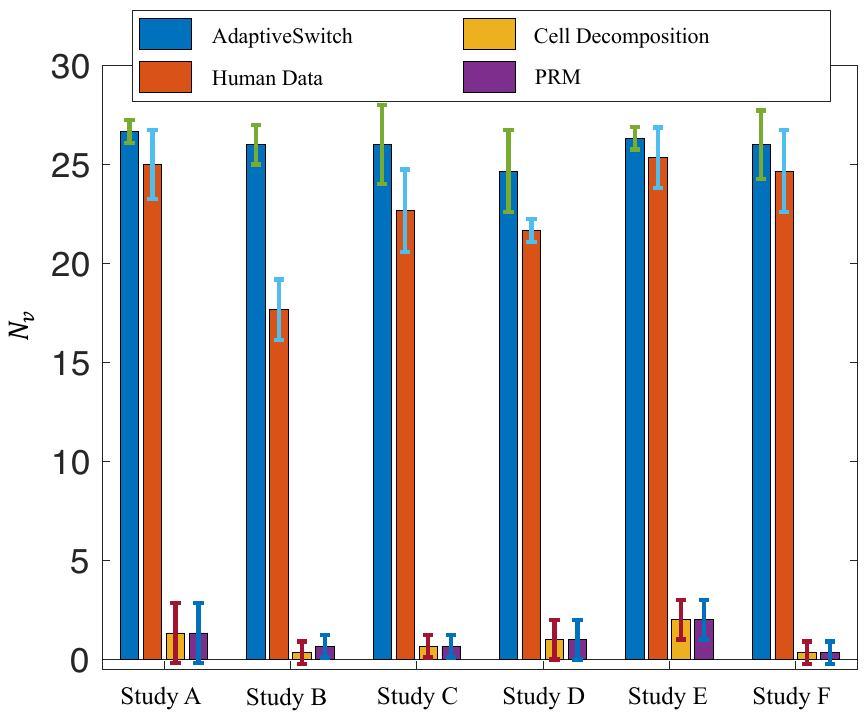}
	}
	\hfill
	\subfloat[\label{subfig:FogHumanD}]{
		\includegraphics[width = 0.45\textwidth]{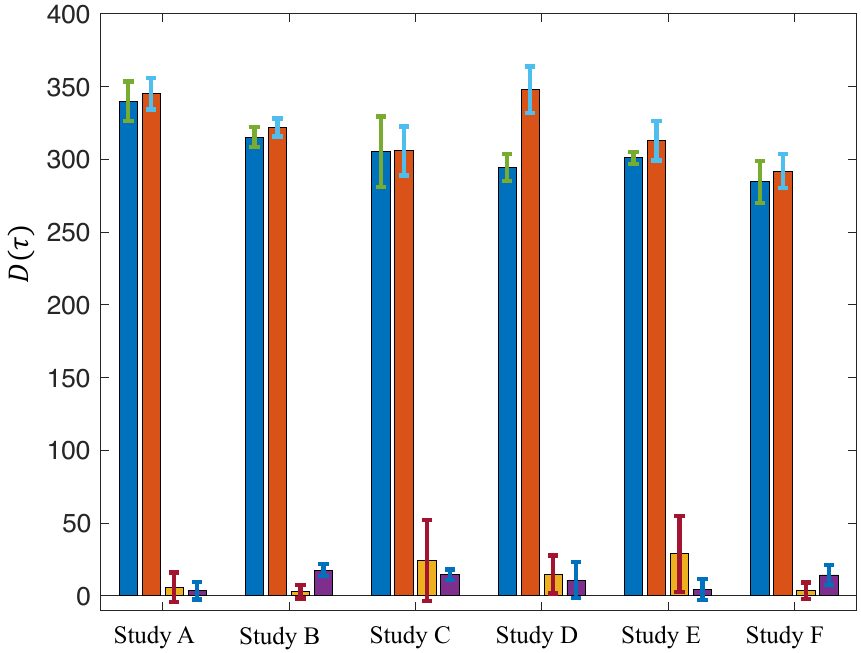}
	}
	\hfill
	\caption{(a) Number of classified targets and (b) travel distance of AdaptiveSwitch
		optimal strategies and the human strategy.}
	\label{fig:HeuristicCompareHumanStudy}
	
\end{figure}

The performance of AdaptiveSwitch is compared with that of optimal strategies and the human
strategy. One important metric used to evaluate a strategy's capability to search for targets in fog conditions
is the number of classified targets: $N_v$. Under fog pressure, as shown in Fig. \ref{subfig:FogHumanNv}
and Fig. \ref{subfig:FogHumanD}, the optimal strategies face difficulties in moving and classifying targets,
because of the lack of prior information on the target and obstacle layouts. In contrast, both the human
strategies and the AdaptiveSwitch are able to explore the unknown environment, and even at times
do not capture target information through $\mathcal{S}_P$. AdaptiveSwitch, in particular, achieves
slightly higher target classification rates and shorter travel distances than the human strategy.

\subsubsection{Extended Performance Tests in Simulations}
\label{subsubsec:AdaptiveSwitchTests}

This paper also presents new workspaces and target layouts beyond those used in the active satisficing
experiment. These new layouts are used to assess the performance of AdaptiveSwitch in different
environments and to determine its applicability beyond the specific experimental settings.

\paragraph{Simulations with Fixed Truncated Sensing Range}
\label{para:MATLABSim}

\begin{figure*}[t]
	\centering
	\subfloat[\label{subfig:SimCaseStudy1}]{
		\includegraphics[width = 0.41\textwidth]{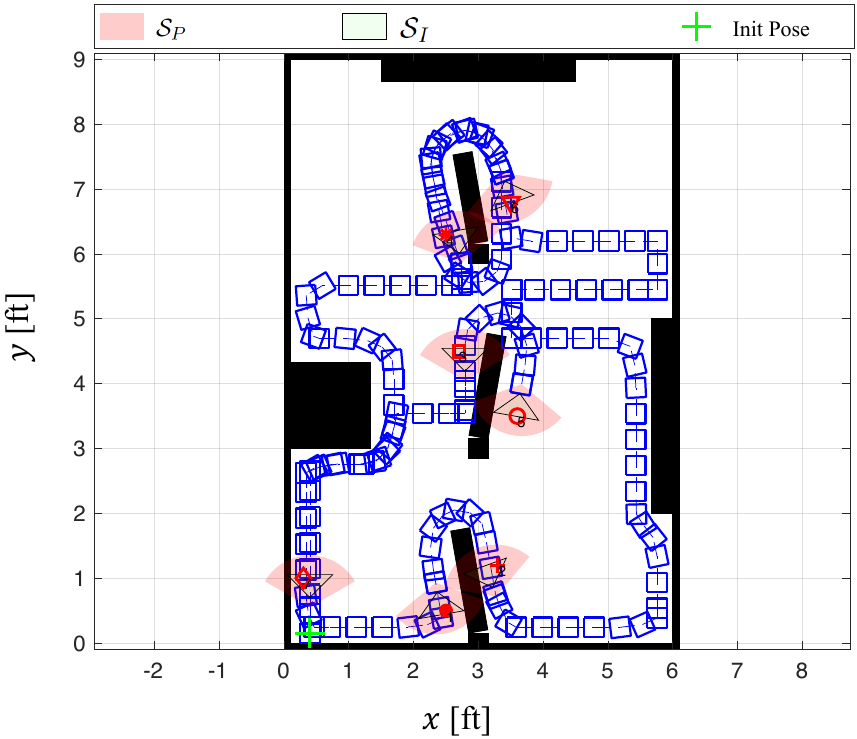}
	}
	\hfill
	\subfloat[\label{subfig:SimCaseStudy2}]{
		\includegraphics[width = 0.41\textwidth]{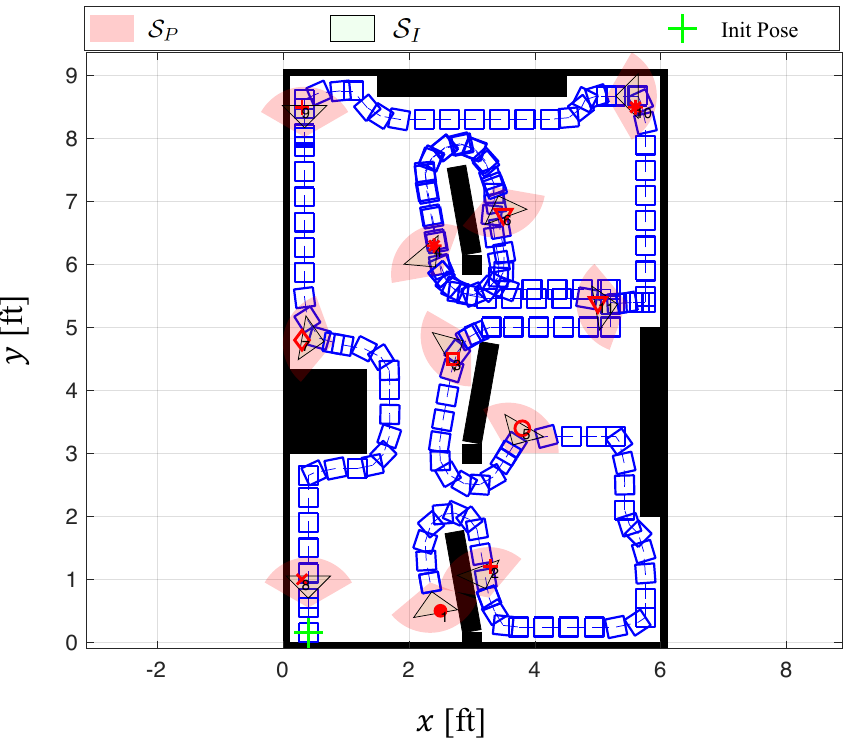}
	}
	\hfill
	\subfloat[\label{subfig:SimCaseStudy3}]{
		\includegraphics[width = 0.41\textwidth]{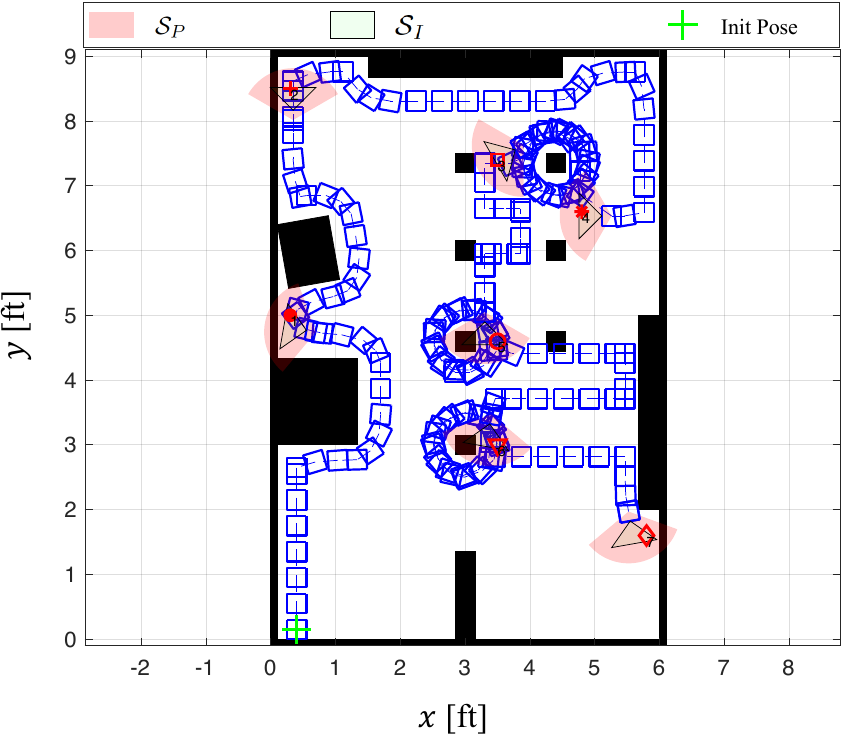}
	}
	\hfill
	\subfloat[\label{subfig:SimCaseStudy4}]{
		\includegraphics[width = 0.41\textwidth]{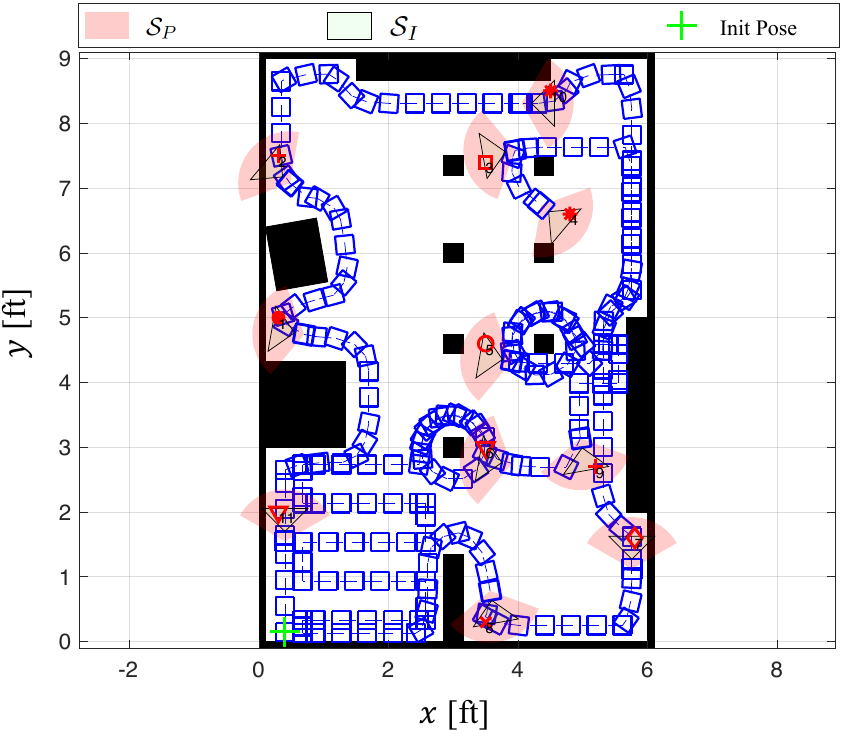}
	}
	\caption{Four workspace in MATLAB\textregistered$\,$ simulations and
		AdaptiveSwitch trajectories for case studies (a)-(d).}
	\label{fig:SimCaseStudies}
\end{figure*}

The evaluation of AdaptiveSwitch involves conducting simulations in MATLAB\textregistered$,$ by
using four newly designed workspaces and corresponding target layouts. The effect of fog is emulated
by imposing a fixed truncated sensing range for $\mathcal{S}_P$, and the trajectory of AdaptiveSwitch
is superimposed on each workspace to observe its behavior (Fig. \ref{fig:SimCaseStudies}). The
simulations consider fixed geometries for the FOV of the onboard sensors, assume no target miss
detection or false alarms, and assume perfect target feature recognition. ForwardExplore
and two optimal strategies are also implemented for comparison.

\begin{figure}[h]
	\centering
	\centering
	\subfloat[\label{subfig:FogMATLABSimPerform1}]{
		\includegraphics[width = 0.48 \textwidth]{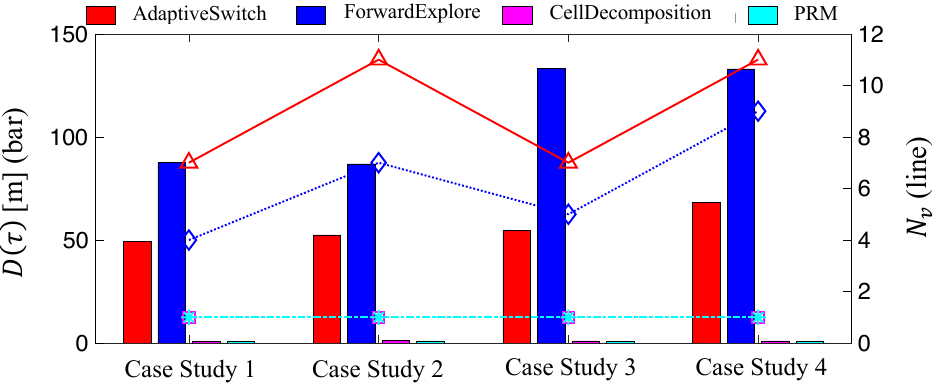}
	}
	\hfil
	\subfloat[\label{subfig:FogMATLABSimPerform2}]{
		\includegraphics[width = 0.48 \textwidth]{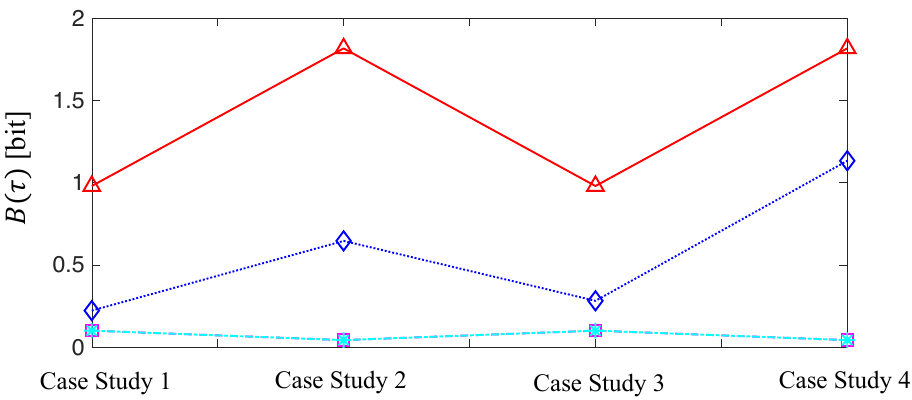}
	}
	\caption{(a). Number classified targets and travel distance (b) information
		gain for two heuristic strategies and two optimal strategies in four case studies.}
	\label{fig:FogMATLABSimPerform}
\end{figure}

The results of the simulations demonstrate that the optimal strategies perform poorly in terms of travel
distance $D(\tau)$ and the number of classified targets ($N_v$) (Fig. \ref{subfig:FogMATLABSimPerform1}),
owing to the challenges posed by fog and limited sensing capabilities. In contrast, the proposed heuristics
(AdaptiveSwitch and ForwardExplore) outperform the optimal strategies in terms of $N_v$, because they
are able to explore the workspace even when no targets were visible.

Additionally, AdaptiveSwitch is more efficient than ForwardExplore in terms of travel distance. By adapting
its exploration strategy and leveraging the combination of three simple heuristics, AdaptiveSwitch
is able to classify more targets while traveling shorter distances. Consequently, higher information value $B(\tau)$
than that with both ForwardExplore and the optimal strategies is observed across all four case studies
(Fig. \ref{subfig:FogMATLABSimPerform2}). These findings highlight the effectiveness of the AdaptiveSwitch in
navigating foggy environments and its superiority to the optimal strategies and the ForwardExplore in
terms of information gathering and travel efficiency.

\paragraph{Simulations with Artificial Fog}
\label{para:WebotsSim}

\begin{figure}[h]
	
	\centering
	\subfloat[\label{subfig:NewWbWS1}]{
		\includegraphics[width = 0.22 \textwidth]{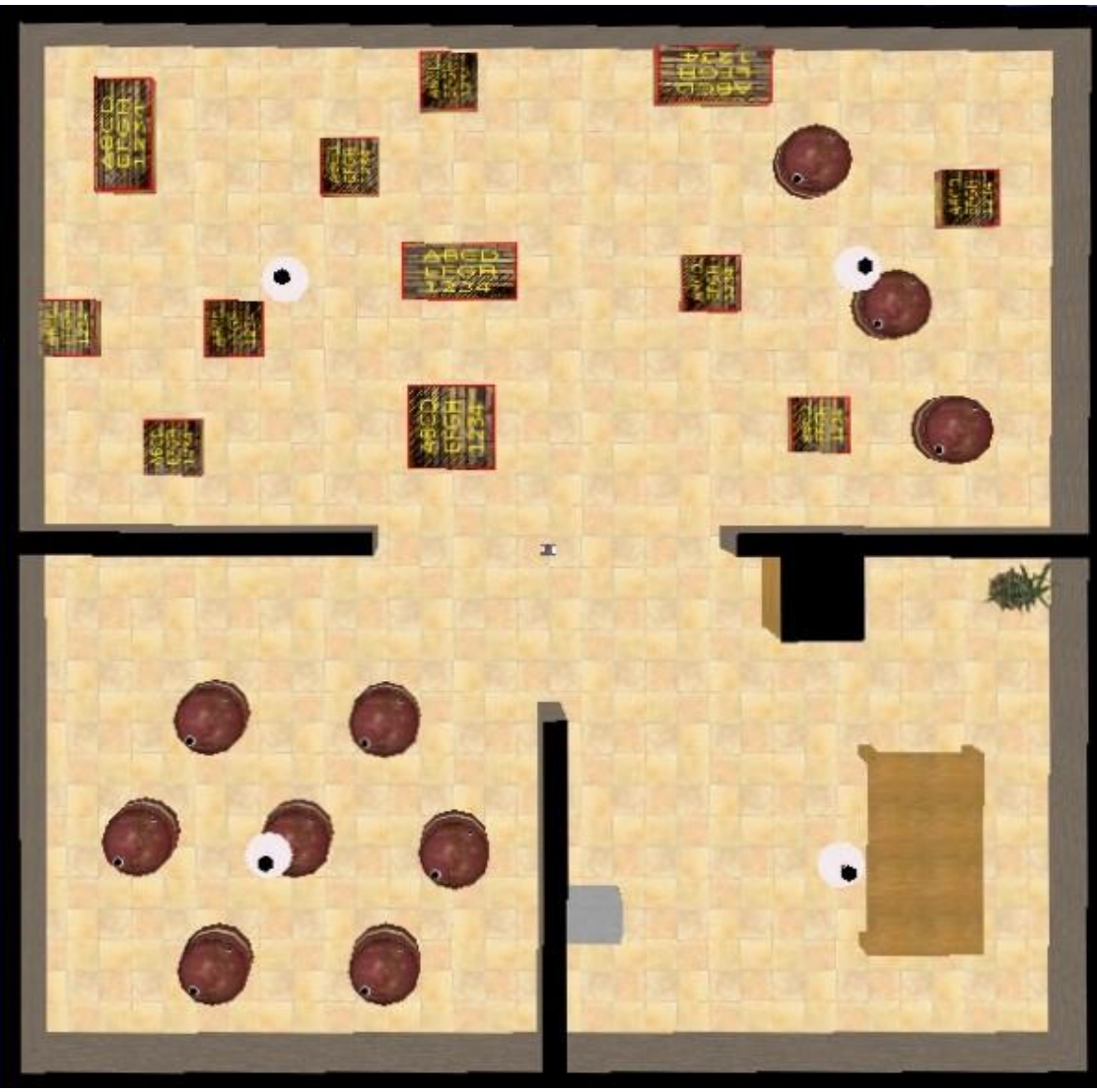}
	}
	\subfloat[\label{subfig:NewWbWS2}]{
		\includegraphics[width = 0.21 \textwidth]{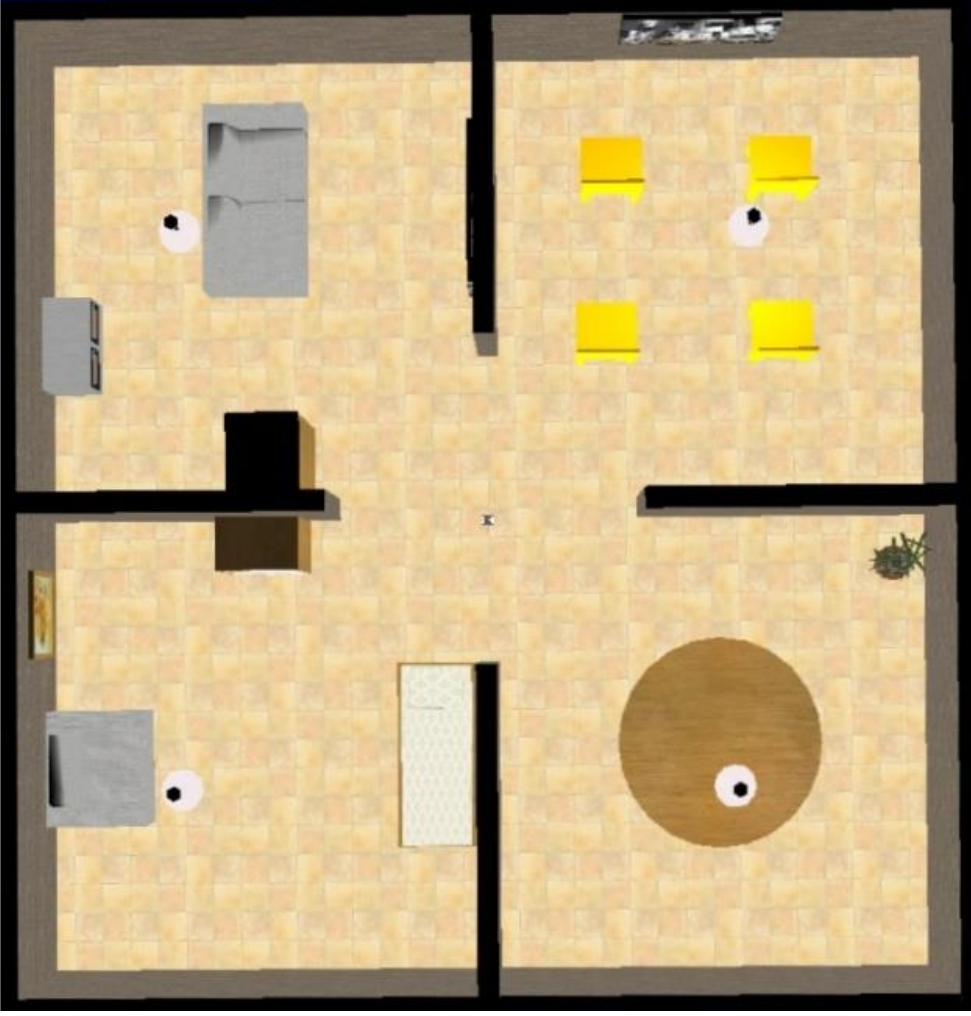}
	}
	
	\caption{New designs of workspace for heuristic strategy tests.}
	\label{fig:WebotsNewWorkspace}
\end{figure}

Two new workspaces are designed in Webots\textregistered$\,$ as shown in
Fig. \ref{fig:WebotsNewWorkspace}. The performance of AdaptiveSwitch and its
standalone heuristics for the two workspaces is shown in Table \ref{tab:WebotsSimWS1}
and Table \ref{tab:WebotsSimWS2}. The comparison reveals the substantial advantage
of AdaptiveSwitch. In both workspace scenarios, as shown in Table \ref{tab:WebotsSimWS1}
and \ref{tab:WebotsSimWS2}, AdaptiveSwitch outperforms its standalone heuristics by
successfully finding and classifying all targets within the given simulation time upper bound.
In contrast, the standalone heuristics are unable to achieve this level of performance.
AdaptiveSwitch not only visits and classifies all targets, but also accomplishes the tasks
within shorter travel distances than the standalone heuristics. Therefore, AdaptiveSwitch
exhibits higher target visitation efficiency ($\eta_v$) which is calculated as the ratio of the
number of classified targets to the travel distance ($N_v/D(\tau)$). The target visitation
efficiency of AdaptiveSwitch is at least twice higher than that of the standalone heuristics.

These results highlight the strength of combination used by AdaptiveSwitch. By integrating
multiple simple heuristics, AdaptiveSwitch demonstrates a greater ability to explore the
entire environment in the presence of fog. In contrast, the standalone heuristics tend to
be less flexible and may become trapped in certain ``moving patterns"; therefore, although
they can explore some areas effectively, they might struggle to reach other areas.

\begin{table*}[h]
	\caption{Performance comparison of AdaptiveSwitch and Standalone
		heuristics in Webots\textregistered$\,$: Workspace A}
	\label{tab:WebotsSimWS1}
	
	\centering
	\begin{tabular}{|p{4.5cm}||p{1.8cm}|p{1.5cm}|p{1.5cm}|}
		\hline
		\multirow{2}{4em}{Performance Metrics} & \multicolumn{3}{|c|}{Heuristic Strategies} \\
		\cline{2-4}
		& AdaptiveSwitch & RandomWalk & AreaCoverage \\
		\hline
		Travel distance, $D(\tau)\,[\text{m}]$  &  \hfil \textbf{86.19} &  \hfil 164.87 & \hfil 224.18 \\
		Number of classified targets, $N_v$ &  \hfil \textbf{7}/7 & \hfil 7/7 & \hfil 3/7\\
		Target visitation efficiency, $\eta_v \, [\text{m}^{-1}]$ &  \hfil \textbf{0.0812} & \hfil 0.0425 & \hfil 0.0134\\
		\hline
		Travel distance, $D(\tau)\,[\text{m}]$ &  \hfil \textbf{148.98} &  \hfil 291.69 & \hfil 246.38 \\
		Number of classified targets, $N_v$ &  \hfil \textbf{13}/13 & \hfil 11/13 & \hfil 6/13\\
		Target visitation efficiency, $\eta_v \, [\text{m}^{-1}]$ &  \hfil \textbf{0.0873} & \hfil 0.0377 & \hfil 0.0244\\
		\hline
		Travel distance, $D(\tau)\,[\text{m}]$ &  \hfil \textbf{159.97} & \hfil 236.86 & \hfil 205.78 \\
		Number of  classified targets, $N_v$ &  \hfil \textbf{15}/15 & \hfil 11/15 & \hfil 8/15\\
		Target visitation efficiency, $\eta_v \, [\text{m}^{-1}]$ &  \hfil \textbf{0.0938} & \hfil 0.0464 & \hfil 0.0389\\
		\hline
	\end{tabular}
	
\end{table*}

\begin{table*}[h]
	\caption{Performance comparison of AdaptiveSwitch and Standalone heuristics in
		Webots\textregistered$\,$: Workspace B}
	\label{tab:WebotsSimWS2}
	\centering
	\begin{tabular}{|p{4.5cm}||p{1.8cm}|p{1.5cm}|p{1.5cm}|}
		\hline
		\multirow{2}{4em}{Performance Metrics} & \multicolumn{3}{|c|}{Heuristic Strategies} \\
		\cline{2-4}
		& AdaptiveSwitch & RandomWalk & AreaCoverage \\
		\hline
		Travel distance, $D(\tau)\,[\text{m}]$ &  \hfil \textbf{122.86} &  \hfil 218.72 & \hfil 265.49 \\
		Number of  classified targets, $N_v$ &  \hfil \textbf{7}/7 & \hfil 5/7 & \hfil 5/7\\
		Target visitation efficiency,  $\eta_v \, [\text{m}^{-1}]$ &  \hfil \textbf{0.0570} & \hfil 0.0229 & \hfil 0.0188\\
		\hline
		Travel distance, $D(\tau)\,[\text{m}]$ &  \hfil \textbf{122.57} &  \hfil 219.49 & \hfil 234.70 \\
		Number of  classified targets, $N_v$ &  \hfil \textbf{13}/13 & \hfil 10/13 & \hfil 7/13\\
		Target visitation efficiency,  $\eta_v \, [\text{m}^{-1}]$ &  \hfil \textbf{0.0873} & \hfil 0.0456 & \hfil 0.0298\\
		\hline
		Travel distance: $D(\tau)\,[\text{m}]$ &  \hfil \textbf{129.19} & \hfil 226.57 & \hfil 216.25 \\
		Number of classified targets, $N_v$ &  \hfil \textbf{15}/15 & \hfil 12/15 & \hfil 8/15\\
		Target visitation efficiency,  $\eta_v \, [\text{m}^{-1}]$ &  \hfil \textbf{0.1161} & \hfil 0.0530 & \hfil 0.0370\\
		\hline
	\end{tabular}
\end{table*}

\subsubsection{Physical Experiment Tests in Real Fog Environment}
\label{subsubsec:Physicalexperiments}

To handle real-world uncertainties that are not adequately modeled in
simulations, this paper conducts physical experiments to test the AdaptiveSwitch.
These uncertainties include factors such as the robot's initial position and orientation,
target miss detection and false alarms, depth measurement errors, and control disturbances.
In addition, the fog models available in Webots\textregistered$,$ are relatively simple and
do not provide a wide range of possibilities for simulating the degrading effects of fog on
target detection and classification performance. Consequently, this paper performs physical
experiments to better capture the complexities and uncertainties associated with real-world
conditions.

The physical experiments use the ROSbot2.0 robot equipped with
an RGB-D camera as the primary sensor. The YOLOv3 object detection algorithm is employed to
detect the targets of interest (e.g., an apple, watermelon, orange, basketball, computer, book,
cardboard box, and wooden box) identical to those in human experiments. The
training images for the YOLOv3 are captured in a clear environment.

\begin{figure}[h]
	\centering
	\subfloat[\label{subfig:SensingInterruptionClear}]{
		\includegraphics[width = 0.23 \textwidth]{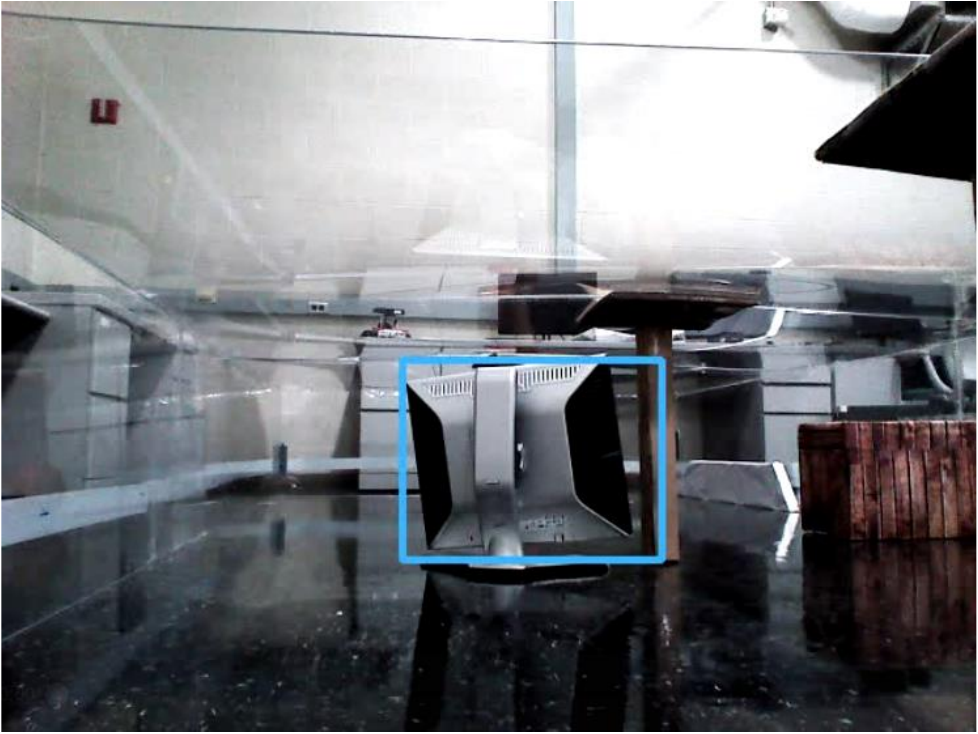}
	}
	\hfil
	\subfloat[\label{subfig:SensingInterruptionFog}]{
		\includegraphics[width = 0.23 \textwidth]{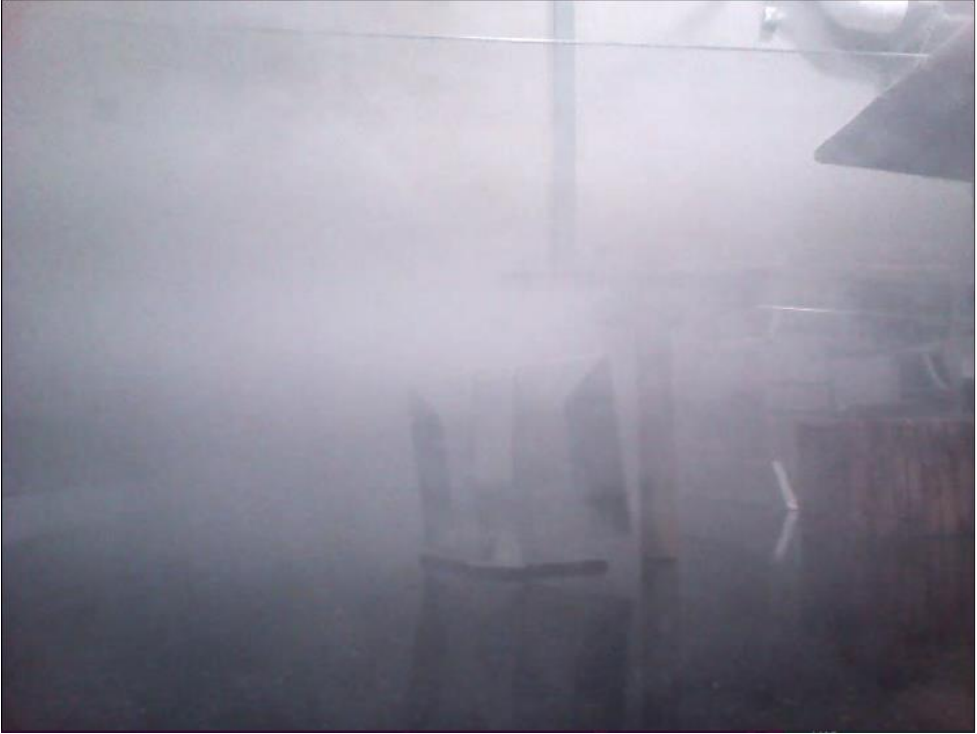}
	}
	\caption{Object detection results (a) in clear and (b) fog conditions.}
	\label{fig:SensingInterruption}
\end{figure}

As depicted in Fig. \ref{fig:SensingInterruption}, the YOLOv3 algorithm successfully
detects the existence of the target ``computer" when the environment is clear, as shown in
Fig. \ref{subfig:SensingInterruptionClear}. However, when fog is present, as illustrated in
Fig. \ref{subfig:SensingInterruptionFog}, the algorithm fails to detect the target. This result
demonstrates the degrading effect on the performance of target detection algorithms.

In the physical experiments conducted with ROSbot2.0 \cite{ROSbotHuarsion18},
AdaptiveSwitch and ForwardExplore are implemented to test their performance in an
environment with fog. A plastic box is constructed with dimensions 10'0" x 6'0" x 1'8" in
order to create the foggy environment. The box is designed to contain different layouts of obstacles
and targets, capturing various aspects of a ``treasure hunt" scenario, such as target density
and target view angles. Each heuristic strategy is tested five times in each layout, considering
all the uncertainties described earlier. The travel distances in the physical experiments are
measured in inertial measurement unit.

\begin{figure}[h]
	\centering
	\subfloat[\label{subfig:PhysicalLayout1Clear}]{
		\includegraphics[width = 0.45 \textwidth]{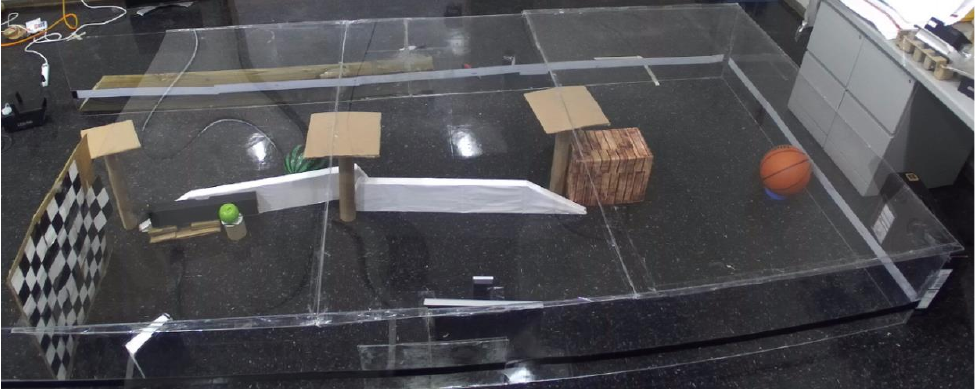}
	}
	\hfill
	\subfloat[\label{subfig:PhysicalLayout1Fog}]{
		\includegraphics[width = 0.45 \textwidth]{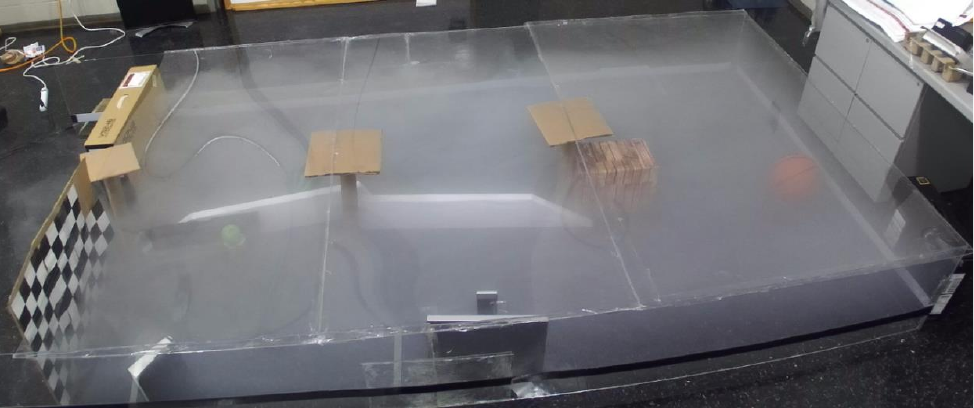}
	}
	\caption{The first workspace and target layout for the physical experiment under
		(a) clear and (b) fog condition.}
	\label{fig:PhysicalExpLayout1}
\end{figure}

\begin{figure}[h]
	\centering
	\includegraphics[width = 0.45 \textwidth]{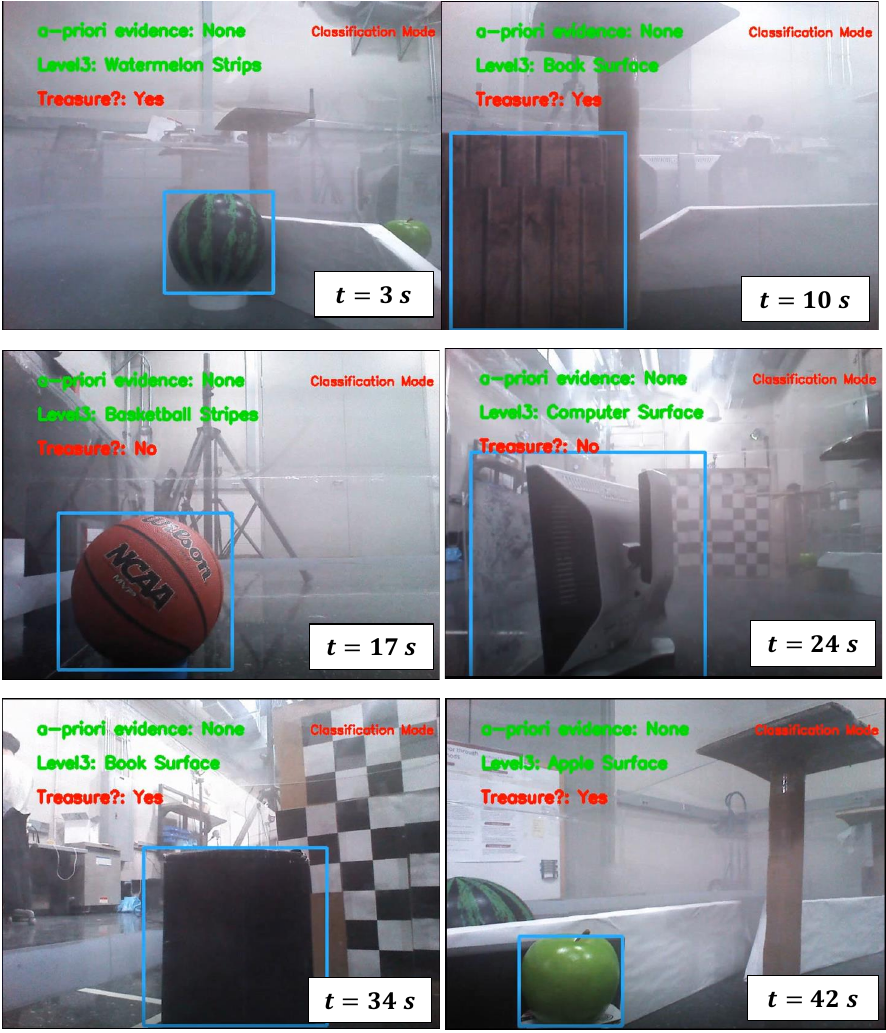}
	\caption{Target visitation sequence of AdaptiveSwitch in the first workspace.}
	\label{fig:ASLayout1}
\end{figure}

The first layout (Fig. \ref{fig:PhysicalExpLayout1}) comprises of six targets: a watermelon,
wooden box, basketball, book, apple, and computer. The target visitation sequences of AdaptiveSwitch
along the path are depicted in Fig. \ref{fig:ASLayout1}, showing the robot's trajectory and the
order in which the targets are visited. The performance of the two strategies is summarized in
Table \ref{tab:performlayout1}, as evaluated according to three aspects: travel distance $D(\tau)$,
correct target feature classifications, and information gathering efficiency $\eta_B$. These metrics
assess the quality of the strategies' action and test decisions.

\begin{table}[h]
	\caption{Performance Comparison of Heuristic Strategies in target layout 1}
	\label{tab:performlayout1}
	\centering
	\begin{tabular}{|p{4.1cm}||p{1.9cm}|p{1.8cm}|}
		\hline
		\multirow{2}{4em}{Performance Metrics} & \multicolumn{2}{|c|}{Heuristic Strategies} \\
		\cline{2-3}
		& AdaptiveSwitch & ForwardExplore\\
		\hline
		Number of classified targets, $N_v$ &  \hfil 6/6 & \hfil 6/6 \\
		Travel distance, $D(\tau)\,[\text{m}]$  &  \hfil \textbf{6.43 $\pm$ 0.90} &  \hfil 8.38 $\pm$ 2.07 \\
		Correct target feature classifications &  \hfil \textbf{13.40 $\pm$ 1.82} &  \hfil 12.40 $\pm$ 1.95 \\
		Info gathering efficiency, $\eta_B \, [\text{bit/m}]$ &  \hfil \textbf{0.155 $\pm$ 0.023} & \hfil 0.090 $\pm$ 0.018\\
		\hline
	\end{tabular}
\end{table}

\begin{figure}[h]
	\centering
	\subfloat[\label{subfig:PhysicalLayout2Clear}]{
		\includegraphics[width = 0.5 \textwidth]{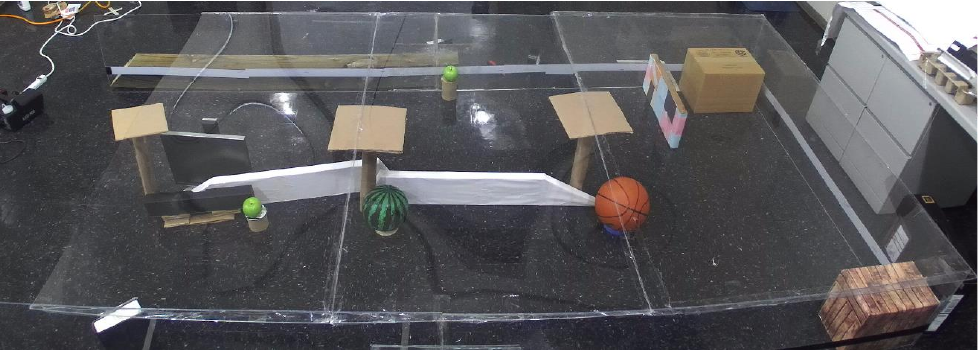}
	}
	\hfill
	\subfloat[\label{subfig:PhysicalLayout2Fog}]{
		\includegraphics[width = 0.5 \textwidth]{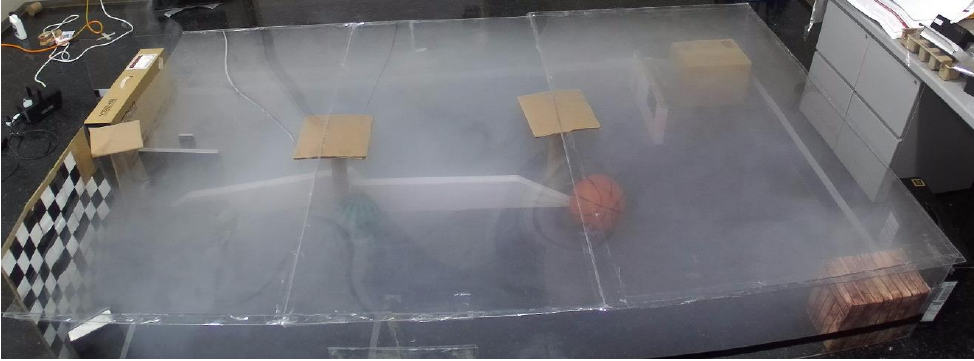}
	}
	\caption{The second workspace and target layout for the physical experiment under
		(a) clear and (b) fog condition.}
	\label{fig:PhysicalExpLayout2}
\end{figure}

\begin{figure}[h]
	\centering
	\includegraphics[width = 0.45 \textwidth]{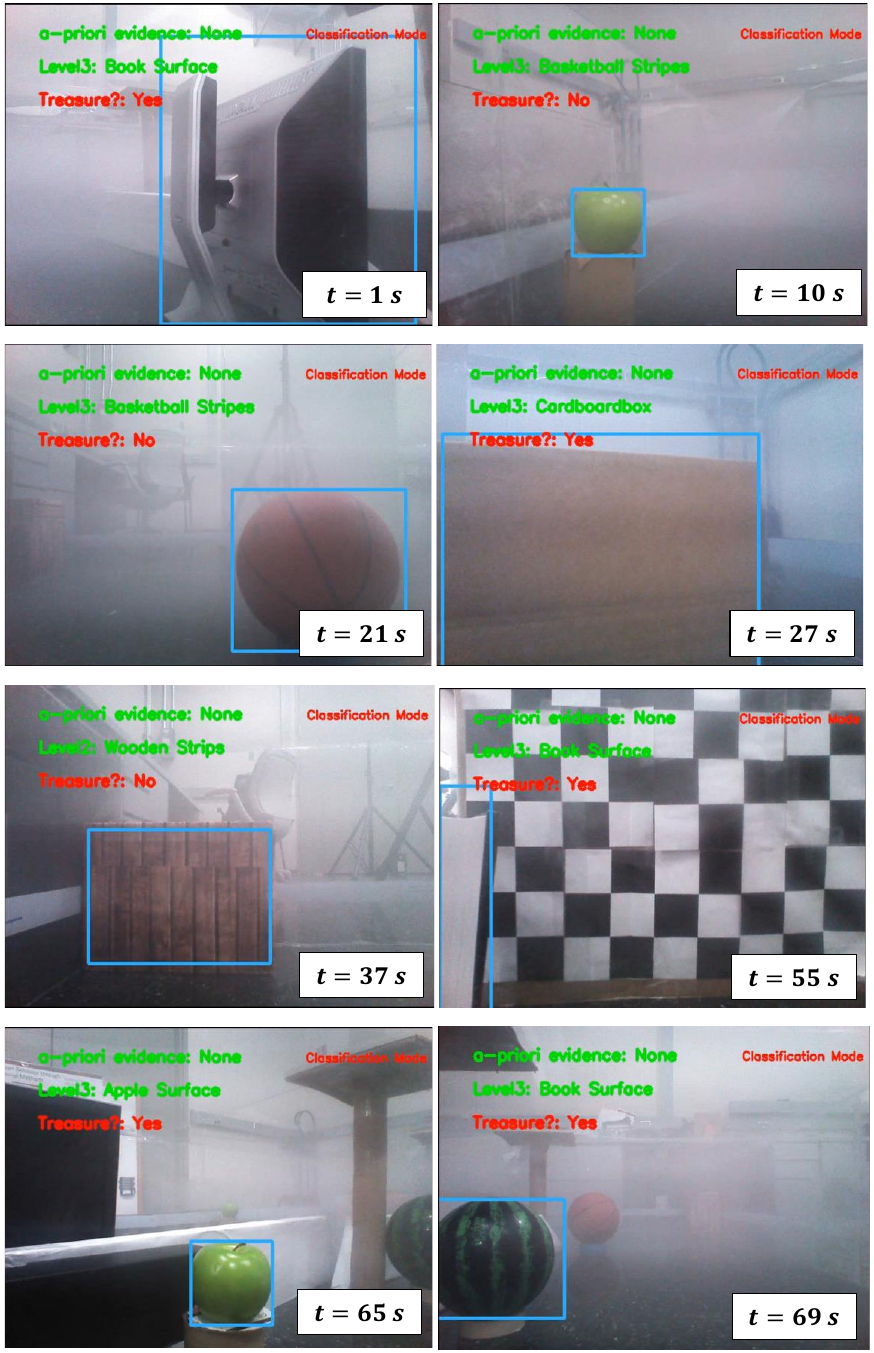}
	\caption{Target visitation sequence of AdaptiveSwitch in the second workspace.}
	\label{fig:ASLayout2}
\end{figure}

The second layout (Fig. \ref{fig:PhysicalExpLayout2}) contains eight targets: a watermelon, wooden box,
basketball, book, computer, cardboard box, and two apples. The obstacles layout is also changed with
respect to the first layout: the cardboard box is placed in a ``corner" and is visible from only one
direction, thus increasing the difficulty of detecting this target. This layout enables a case study in which
the targets are more crowded than in the first layout. The mobile robot first-person-views of AdaptiveSwitch
along the path are demonstrated in Fig. \ref{fig:ASLayout2}, and the performance is shown in
Table \ref{tab:performlayout2}.

\begin{table}[h]
	\caption{Performance Comparison of Heuristic Strategies in target layout 2}
	\label{tab:performlayout2}
	\centering
	\begin{tabular}{|p{4.1cm}||p{1.9cm}|p{1.8cm}|}
		\hline
		\multirow{2}{4em}{Performance Metrics} & \multicolumn{2}{|c|}{Heuristic Strategies} \\
		\cline{2-3}
		& AdaptiveSwitch & ForwardExplore\\
		\hline
		Number of classified targets, $N_v$ &  \hfil 8/8 & \hfil 8/8 \\
		Travel distance, $D(\tau)\,[\text{m}]$  &  \hfil \textbf{8.41 $\pm$ 0.46}  &  \hfil 13.45 $\pm$ 2.10 \\
		Correct target feature classifications &  \hfil \textbf{17.80 $\pm$ 1.10} &  \hfil 15.20 $\pm$ 1.64 \\
		Info gathering efficiency, $\eta_B \, [\text{bit/m}]$ &  \hfil \textbf{0.151 $\pm$ 0.008} & \hfil 0.091 $\pm$ 0.016\\
		\hline
	\end{tabular}
\end{table}

\begin{figure}[h]
	\centering
	\subfloat[\label{subfig:PhysicalLayout3Clear}]{
		\includegraphics[width = 0.45 \textwidth]{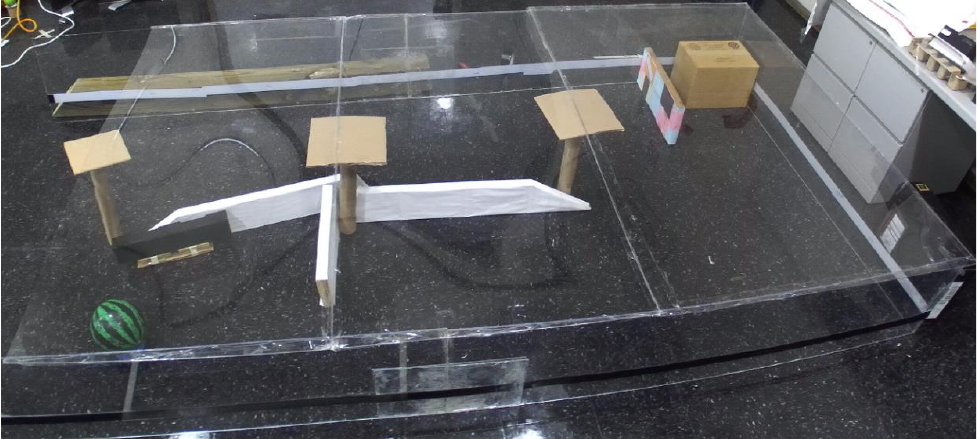}
	}
	\hfill
	\subfloat[\label{subfig:PhysicalLayout3Fog}]{
		\includegraphics[width = 0.45 \textwidth]{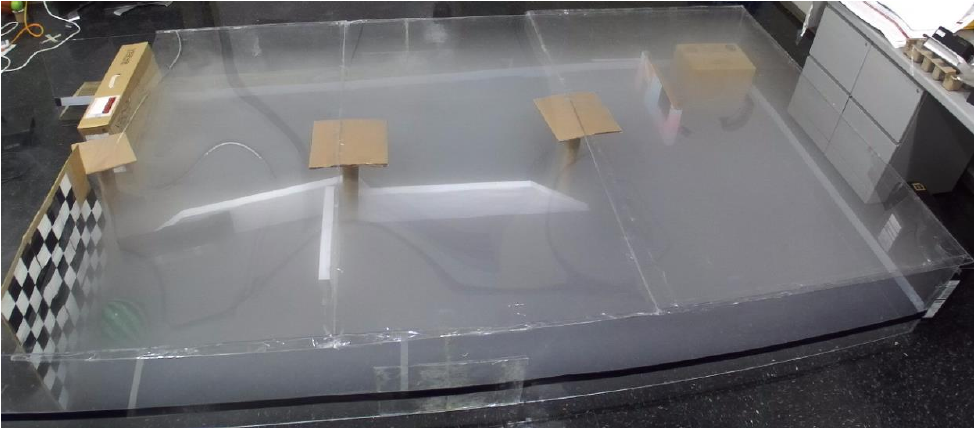}
	}
	\caption{The third workspace and target layout for the physical experiment under
		(a) clear and (b) fog condition.}
	\label{fig:PhysicalExpLayout3}
\end{figure}

\begin{figure}[h]
	\centering
	\includegraphics[width = 0.45 \textwidth]{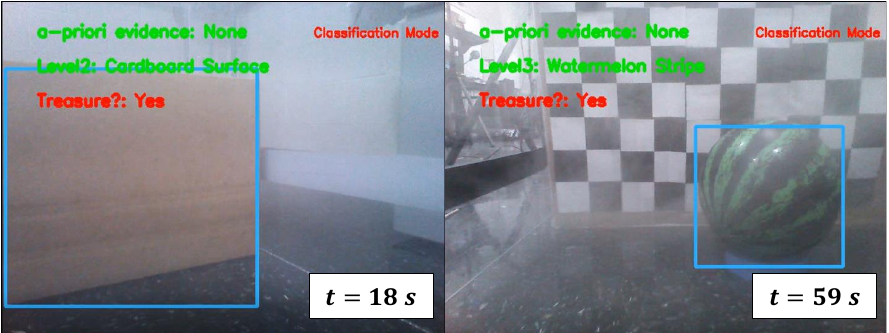}
	\caption{Target visitation sequence of AdaptiveSwitch in the third workspace.}
	\label{fig:ASLayout3}
\end{figure}

The third layout (Fig. \ref{fig:PhysicalExpLayout3}) contains two targets: a cardboard box,
and a watermelon. Note that having fewer targets does not necessarily make the problem easier, because
the difficulty in target search in fog comes from how to navigate when no target is in the FOV.
This layout intentionally makes the problem ``difficult", because it ``hides" two targets
behind the walls. The mobile robot first-person-views of AdaptiveSwitch along the path are
demonstrated in Fig. \ref{fig:ASLayout3}, and the performance is shown in Table \ref{tab:performlayout3}.
The videos for all physical experiments (AdaptiveSwitch and ForwardExplore in three layouts)
are accessible through the link in \cite{ChenUGVFOGLISCExperiments22}.

\begin{table}[h]
	\caption{Performance Comparison of Heuristic Strategies in target layout 3}
	\label{tab:performlayout3}
	\centering
	\begin{tabular}{|p{4.1cm}||p{1.8cm}|p{1.8cm}|}
		\hline
		\multirow{2}{4em}{Performance Metrics} & \multicolumn{2}{|c|}{Heuristic Strategies} \\
		\cline{2-3}
		& AdaptiveSwitch & ForwardExplore\\
		\hline
		Number of classified targets, $N_v$ &  \hfil 2/2 & \hfil 2/2 \\
		Travel distance, $D(\tau)\,[\text{m}]$  &  \hfil \textbf{7.48 $\pm$ 0.465} &  \hfil 11.67 $\pm$ 1.37 \\
		Correct target feature classifications &  \hfil \textbf{5.00 $\pm$ 1.00} &  \hfil 4.80 $\pm$ 1.64 \\
		Info gathering efficiency, $\eta_B \, [\text{bit/m}]$ &  \hfil \textbf{0.033 $\pm$ 0.003} & \hfil 0.021 $\pm$ 0.002\\
		\hline
	\end{tabular}
\end{table}

According to the performance summaries in Table \ref{tab:performlayout1}, Table \ref{tab:performlayout2},
and Table \ref{tab:performlayout3}, both AdaptiveSwitch and ForwardExplore are capable of visiting
and classifying all targets in the three layouts under real-world uncertainties. However, AdaptiveSwitch
demonstrates several advantages over ForwardExplore:

\begin{enumerate}
	
	\item  The average travel distance of AdaptiveSwitch is \textbf{30.33\%}, \textbf{59.93\%}, and
	\textbf{56.02\%} more efficient than ForwardExplore in the three workspaces, respectively. This 
	finding indicates that AdaptiveSwitch is able to search target with a shorter travel distance than 
	ForwardExplore.
	
	\item The target feature classification performance of AdaptiveSwitch is slightly better than that 
	of ForwardExplore, with improvements of 8.06\%, 17.11\%, and 4.16\% in the three workspace, 
	respectively. One possible explanation for these results is that the ``obstacle follow" and ``area 
	coverage" heuristics in AdaptiveSwitch cause the robot's body to be parallel to obstacles during 
	classification of target features, thus ensuring that the targets are the major part of the robot's 
	first-person view and make them relatively easier to classify. In contrast, ForwardExplore does 
	not always lead the robot body to be parallel to obstacles during classification, thereby sometimes 
	allowing obstacles to dominate the robot's first-person view and decreasing the target classification 
	performance.
	
\end{enumerate}

\section{Summary and Conclusion}
\label{sec:Conclusion}

This paper presents novel satisficing solutions that modulate between near-optimal and heuristics 
to solve satisficing treasure hunt problem under environment pressures. These proposed solutions 
are derived from human decision data collected through both passive and active satisficing 
experiments. The ultimate goal is to apply these satisficing solutions to autonomous robots. The 
modeled passive satisficing strategies adaptively select target features to be entered in measurement 
model based on a given time pressure. The idea behind this approach is the human participants 
behavior that dropping less informative features for inference in order to meet the decision deadline. 
The results show that the modeled passive satisficing strategies outperform the ``optimal" strategy 
that always use all available features for inference in terms of classification performance and 
significantly reduce the complexity of target feature search compared with exhaustive search.

Regarding the active satisficing strategies, the strategy that deals with information cost formulates 
an optimization problem with the hard constraint imposed by information cost. This approach is 
taken because the information cost constraint doesn't fundamentally undermine the accuracy of 
the model of the world and the agent, and optimization still yield high-quality decisions.
The results show that the strategy outperforms human participants across several key metrics 
(e.g., travel distance and measurement productivity, etc.). However, under sensory deprivation, 
the knowledge of the world is severely compromised, and thus decisions produced by optimization 
is risky or even no longer feasible, which is also demonstrated through experiments in this paper. 
The modeled human strategies named AdaptiveSwitch shows the ability to use local information 
and navigate in foggy environment by using heuristics derived from humans. The results also 
show that the AdaptiveSwitch can adapt to varying workspaces with different obstacle layouts, 
target density, etc., beyond the workspace used in the active satisficing experiments. Finally, 
AdaptiveSwitch is implemented on a physical robot and conducts satisificing treasure hunt 
with actual fog, which demonstrates the ability to deal with real-life uncertainties in both 
perception and action.

Overall, the proposed satisficing strategies comprise of a toolbox, which can be readily 
deployed on a robot in order to address different real-life environment pressures encountered 
during the mission. These strategies provide solutions to scenarios characterized by  time 
limitations, constraints on available resources (e.g., fuel or energy), and adverse weathers such as
fog or heavy rain.

\appendices
\section{Mathematical Properties of Heuristics Under Time Pressure}
\label{app:TPMathproperties}

\subsection{Discounted Cumulative Probability Gain (ProbGain)}

\subparagraph{\textbf{Proposition} 1.}
A sufficient condition for \textit{ProbGain} to use all $\mathit{p}$ features is that
the allowable time $t_T$ to make a decision satisfies:

\begin{equation}
    \label{eq:proposition1}
    t_T \geq \frac{\lambda \mathit{p}}{\ln(1 + \frac{\alpha}{\mathit{p}})}
\end{equation}

\noindent
where $\alpha = v_I(x_{\mathit{p}})/v_I(x_{1})$ is the ratio of information
values between the least informative feature and the most informative feature.

\subparagraph{\textbf{Proposition} 2.}
A sufficient condition for \textit{ProbGain} to use 1 (the least possible
number of features to use) feature is that the allowable time to make a
decision $t_T$ satisfies

\begin{equation}
    \label{eq:proposition2}
    t_T \leq \frac{\lambda}{\ln(\mathit{p})}
\end{equation}

\subparagraph{\textbf{Proposition} 3.}

Monotonicity with respect to allowable time $t_T$ for a classification
task with features $\{x_i\}_{i=1}^\mathit{p}$, $H_{\text{ProbGain}}(t_T,\{x_i\}_{i=1}^\mathit{p})$
satisfies:

\begin{equation}
	\label{eq:probposition3}
	H_{\text{ProbGain}}(t_{T,2},\{x_i\}_{i=1}^\mathit{p}) \geq H_{\text{ProbGain}}(t_{T,1},\{x_i\}_{i=1}^\mathit{p})
\end{equation}

\noindent
for  $\forall t_{T,1}, t_{T,2},\,\,t_{T,2} > t_{T,1}$

Propositions 1 and 2 indicate the behavior of \text{ProbGain} under
``extreme" conditions. Notably, proposition 1 shows that as
the allowable time $t_T \geq \frac{\lambda \mathit{p}}{\ln(1 + \frac{\alpha}{\mathit{p}})}$,
the heuristic uses all features to make the classification decision
(i.e., converges to the ``optimal strategy," which uses all features to make
a decision). In addition, according to Proposition 2, when the allowable
time is too short ($t_T \leq \frac{\lambda}{\ln(\mathit{p})}$), the heuristic
only uses one feature (the least possible number of features to use) to make the decision.
Proposition 3 shows the monotonicity of the heuristic with respect to allowable
time $t_T$; as the allowable time increases, the heuristic uses monotonically more
features to make a classification decision.

\subsection{Discounted Log-odds Ratio (LogOdds)}
This heuristic regards the log-odds ratio,

\begin{equation*}
  c_i = log\frac{p(Y = y_1~|~x_{1},...,x_{i})}{p(Y = y_2~|~x_{1},...,x_{i})}
\end{equation*}

\noindent
on the basis of features in set $x_{1},x_{2},,...,x_{i}$ represents the ``confidence" 
of making the classification task. The greater is the value of $|c_i|$, the more 
confident is the classification decision. While one feature comes into consideration, 
an additional time-pressure dependent discount factor is imposed on the absolute 
value the log-odds ratio $c_i$ of the features in set $\{x_{1},x_{2},,...,x_{i}\}$.  
The heuristic selects the features under pressure according to the maximization 
of the product of the discount factors and the log-odds ratio. In this way, less 
informative features are dropped because of the discount factor. As
the time pressure increases, the heuristic has a greater tendency to drop the features.

\subparagraph{\textbf{Proposition} 4.}
A sufficient condition for \textit{LogOdds} to use one feature is if the allowable
time $t_T$ to make a decision satisfies

\begin{equation}
	\label{eq:proposition4}
    t_T \leq \frac{\lambda}{\ln(1 + \frac{\mathit{p}-1}{~|~1 + \beta~|~})}
\end{equation}

\noindent
where $\beta = v_0/v_I(x_1)$.

\subparagraph{\textbf{Proposition} 5.}

Monotonicity with respect to allowable time $t_T$:
for an object with features $\{x_i\}_{i=1}^\mathit{p}$, $H_{\text{LogOdds}}(t_T,\{x_i\}_{i=1}^\mathit{p})$ satisfies:

\begin{equation}
	\label{eq:proposition5}
    H_{\text{LogOdds}}(t_{T,2},\{x_i\}_{i=1}^\mathit{p}) \geq H_{\text{LogOdds}}(t_{T,1},\{x_i\}_{i=1}^\mathit{p})
\end{equation}

\noindent
for $\forall t_{T,1}, t_{T,2},\,\,t_{T,2} > t_{T,1}$.

Note that unlike $H_{\text{ProbGain}}$, although $H_{\text{LogOdds}}$
tends to use more features as time pressure is released, $H_{\text{LogOdds}}$
does not necessarily use all $\mathit{p}$ features when the time available $t_T$ is greater than
a certain threshold, because the value metric used in $H_{\text{LogOdds}}$: $|c_i|
= |v_0 + \sum_{j=1}^i v_I(x_{j})|$ is not monotonically increasing as the number of
features to use $i$ increases.

\subsection{Information Free Feature Number Discounting (InfoFree)}

After sorting the features in terms of the information value, the cut-off criterion of
this heuristic is no longer dependent on the information value. Thus the allowable
decision time $t_T$ is the only argument for the heuristic. As
$\exp(-\frac{\lambda}{t_T}) < 1, t_T > 0$, the number of features to use is
always less than or equal to $M$ and decreases exponentially when time pressure
increases, and the parameter $\lambda > 0$ controls how much a time pressure is discounted.
Given the monotonicity of the exponential function, $H_{\text{InfoFree}}$ uses more
features as time pressure is released and it uses all $\mathit{p}$ features if the time available $t_T$
is sufficiently large, and uses one feature if the time available $t_T$ is sufficiently small.

\section*{Acknowledgments}
This research is funded by the Office of Naval Research Grants
N00014-13-1-0561.

\
\bibliographystyle{IEEEtran}
\bibliography{Yucheng_refs.bib}

\end{document}